\documentclass{article}
\PassOptionsToPackage{dvipsnames,table}{xcolor}

\def\COLMmode{final}

\ifdefined\COLMmode
  \usepackage[\COLMmode]{colm2026_conference}
\else
  \usepackage[submission]{colm2026_conference}
\fi

\usepackage{microtype}
\usepackage{graphicx}
\usepackage{booktabs}
\usepackage{hyperref}
\usepackage{amsmath}
\usepackage{caption}

\usepackage[utf8]{inputenc}
\usepackage[T1]{fontenc}
\usepackage{placeins}    
\usepackage{url}
\usepackage{xurl}
\usepackage{amsfonts}
\usepackage{nicefrac}
\usepackage{xcolor}
\usepackage{tikz}
\definecolor{MLBlue}{HTML}{2F6FA3}
\definecolor{MLVermillion}{HTML}{A84F2A}
\definecolor{MLTeal}{HTML}{1C7A6A}
\definecolor{MLAmber}{HTML}{9A6200}
\definecolor{MLPurple}{HTML}{71579B}
\definecolor{MLGray}{HTML}{5F6368}
\definecolor{MLLightGray}{HTML}{F6F7F8}
\definecolor{MLBorderGray}{HTML}{C9CED6}
\definecolor{MLGreen}{HTML}{2E7D4F}
\definecolor{MLRed}{HTML}{A23B3B}
\tikzset{
  mlpv/role/.style={draw=MLBlue!70!black, fill=MLBlue!10, text=black},
  mlpv/model/.style={draw=MLBlue!70!black, fill=MLBlue!10, text=black},
  mlpv/assistant/.style={draw=MLVermillion!70!black, fill=MLVermillion!10, text=black},
  mlpv/baseline/.style={draw=MLGray!70!black, fill=MLGray!10, text=black},
  mlpv/reference/.style={draw=MLGray!70!black, fill=MLGray!10, text=black},
  mlpv/screen/.style={draw=MLTeal!70!black, fill=MLTeal!10, text=black},
  mlpv/safe/.style={draw=MLTeal!70!black, fill=MLTeal!10, text=black},
  mlpv/saturation/.style={draw=MLAmber!70!black, fill=MLAmber!10, text=black},
  mlpv/risk/.style={draw=MLAmber!70!black, fill=MLAmber!10, text=black},
  mlpv/geometry/.style={draw=MLPurple!70!black, fill=MLPurple!10, text=black},
  mlpv/eval/.style={draw=MLPurple!70!black, fill=MLPurple!10, text=black},
  mlpv/neutral/.style={draw=MLGray!70!black, fill=MLGray!10, text=black},
  mlpv/data/.style={draw=MLTeal!70!black, fill=MLTeal!10, text=black},
}

\usepackage{array}
\usepackage{enumitem}
\usepackage{lineno}      

\newif\iffrp
\frpfalse

\usepackage{xspace}      
\usepackage{xcolor}      
\usepackage{pifont}      
\usepackage{enumitem}    

\newcommand{\refsec}[1]{Section~\ref{#1}}

\newcommand{\reffig}[1]{Figure~\ref{#1}}

\definecolor{darkblue}{rgb}{0, 0, 0.5}
\hypersetup{colorlinks=true, citecolor=darkblue, linkcolor=darkblue, urlcolor=darkblue}
\hypersetup{
  pdftitle={Role Steering of Language Models for Social Simulations},
  pdfauthor={Isaac Song, Mohammed Rehan Parwani, Glenn Matlin, Emile Anand,
    Akhil Theerthala, Arjun Chatterjee, Anthony Wen-Ming Zang, Maria Kostylew,
    Yonadav G. Shavit, Sebastien Krier, Mark Riedl}}

\title{Role Steering of Language Models for Social Simulations}

\usepackage{twemojis}
\newcommand{\AffGT}{\twemoji{bee}}                
\newcommand{\AffMATS}{\twemoji{fire}}             
\newcommand{\AffUIUC}{\twemoji{tangerine}}     
\newcommand{\AffOxford}{\twemoji{open book}}   
\newcommand{\AffDeepMind}{\twemoji{brain}}     
\newcommand{\AffOpenAI}{\twemoji{cyclone}}     
\newcommand{\AffIndep}{\twemoji{compass}}      

\newcommand{\blfootnote}[1]{%
  \begingroup
    \renewcommand{\thefootnote}{}%
    \footnote{#1}%
    \addtocounter{footnote}{-1}%
  \endgroup
}

\author{
  Isaac Song$^{\AffGT\,*}$ \quad
  Mohammed Rehan Parwani$^{\AffGT\,*}$ \quad
  Glenn Matlin$^{\AffGT\,\AffMATS\,*}$ \\
  Emile Anand$^{\AffGT\,\dagger}$ \quad
  Akhil Theerthala$^{\AffIndep\,\dagger}$ \quad
  Arjun Chatterjee$^{\AffUIUC\,\ddagger}$ \\
  Anthony Wen-Ming Zang$^{\AffGT\,\ddagger}$ \quad
  Maria Kostylew$^{\AffMATS\,\AffOxford}$ \quad
  Yonadav G. Shavit$^{\AffOpenAI}$ \\
  Sebastien Krier$^{\AffDeepMind}$ \quad
  Mark Riedl$^{\AffGT}$
  \\[4pt]
  \AffGT\,Georgia Institute of Technology \quad
  \AffMATS\,ML Alignment \& Theory Scholars (MATS) \\
  \AffUIUC\,University of Illinois Urbana-Champaign \quad
  \AffOxford\,University of Oxford \\
  \AffDeepMind\,Google DeepMind \quad
  \AffOpenAI\,OpenAI \quad
  \AffIndep\,Independent
}

\begin{document}

\ifcolmsubmission
\linenumbers
\fi

\maketitle

\ifcolmfinal
  \lhead{Published at the Social Sim'26 Workshop, COLM 2026}
\fi

\blfootnote{$^{*}$Equal contribution. $^{\dagger}$Major contribution.
  $^{\ddagger}$Additional contribution.
  Corresponding author: \texttt{gmatlin@gatech.edu}.}

\begin{abstract}
Social simulations built from language-model agents need role-conditioned
behavior that can be checked before agents are placed into a simulated
population. We introduce an activation-steering screening workflow for
role-conditioned agents: define a role profile, extract a role-specific
direction, sweep four steering coefficients, evaluate role-profile alignment,
and pass or flag each candidate configuration. On OLMo-3-7B-Instruct, we apply
the workflow to a mixed 275-role inventory with 228 role-agnostic questions,
GPT-4.1-mini prompted role references, and GPT-4.1-mini judges. We also introduce CastVectors, which
receive higher judged role-profile alignment than an assistant-axis
directional control from prior persona-vector work, with mean overall scores
of 63.2 versus 41.1 across the tested grid. They also preserve high lexical
diversity, while the control drops sharply at larger coefficients. The
role-level screen is the main contribution: most roles improve as
steering increases, but 38 roles decline across all six measured dimensions,
showing why simulation builders should choose coefficients per role rather
than deploy a uniform high-strength setting. Our code is available at
\url{https://github.com/HCAI-Lab-GT/casting-call-vectors}, and the role vectors
and evaluation data at
\url{https://huggingface.co/datasets/HCAI-Lab-GT/casting-call-vectors}.

\end{abstract}

\section{Introduction}
\label{sec:intro}

Social simulations built from language-model agents inherit the assumptions
encoded in those agents. A workplace, civic process, platform community, or
institutional workflow may require agents that express distinct priorities,
registers, social stances, and decision styles. Prompting a model to ``act
as'' a role is convenient, but small prompt changes can alter behavior,
instruction adherence can drift, and simulated populations can flatten or
misportray heterogeneity
\citep{li_measuring_controlling_2024,lutz_prompt_makes_2025,tosato_persistent_instability_2026}. For social simulation, this creates an attribution problem: an observed interaction pattern may reflect the intended role population, the wording of the role prompts, or shared biases of the underlying model.\looseness=-1

We study activation steering as a way to make role construction measurable
before agents are used in a simulation. Rather than repeatedly specifying a
role in natural language, we extract a direction from the model's activation
space and add it during generation. This intervention exposes a scalar
coefficient that can be varied while holding the model, question, and decoding
setup fixed. Prior work shows that persona- and role-related directions can be
extracted and used to modulate behavior
\citep{chen_persona_vectors_2025,lu_assistant_axis_2026,poterti_can_role_2025,bas_what_can_2026}. In that literature, the Assistant Axis of \citet{lu_assistant_axis_2026} is a reference direction associated with default assistant-like behavior, not with any target role in our inventory. As a comparison condition we use an assistant-axis directional control: for each role, the vector produced by the extraction pipeline of \citet{lu_assistant_axis_2026}, which is near-parallel across roles, evaluated with the same questions and coefficient grid.
The methodological question is how to improve upon the language model's alignment of a role to its true character: define many roles, extract candidate directions, evaluate them consistently, choose coefficients, and identify roles for which stronger steering is counterproductive.

We address this question with a mixed 275-role inventory on
OLMo-3-7B-Instruct \citep{team_olmo_olmo_3_2026}. For each role, we construct a structured profile, generate role-specific elicitation prompts, extract a judge-filtered mean-difference
direction at layer 16, and finally evaluate the resulting candidate agent steered by CastVectors at
$\alpha\in\{1.0,1.5,2.0,2.5\}$ on 228 role-agnostic questions. The principal
comparison uses that assistant-axis control vector. The control is not
scale matched: its vectors have mean $\ell_2$ norm 9.68, compared with 3.79
for the CastVectors, so equal coefficients do not imply equal
perturbation magnitudes.

The resulting picture is heterogeneous rather than uniformly positive.
CastVectors achieve higher mean judged role-profile alignment than
the assistant-axis directional control across the tested grid. For most roles,
alignment rises as $\alpha$ increases: the median per-role correlation is
$r{=}+0.98$, and 74\% of roles improve at every consecutive step. A distinct
minority behaves differently. Thirty-eight roles decline with increasing
$\alpha$ on all six measured dimensions, so they should be flagged rather than
deployed under a uniform high-strength setting. Many of these roles already score highly at the lowest tested coefficient, indicating that the baseline assistant naturally steers strongly toward such roles, in turn making it a coefficient-selection problem rather than a negative reflection of the CastVectors. We call this
category \emph{anti-controllable over the tested range}. Misaligned directional vectors and
over-steering are plausible interpretations, but the current design leaves the
mechanism open.\looseness=-1

\textbf{Scope of the measured outcome.}
Throughout this paper, \emph{role-profile alignment} denotes judged agreement
with the constructed role description and prompted role reference used by our
evaluation pipeline. It measures whether a candidate synthetic agent steered by the CastVectors expresses
the intended profile under the tested questions and coefficients.
\refsec{sec:discussion} states the corresponding validity boundaries and
responsible-use constraints.

\paragraph{Contributions.}
\begin{itemize}[leftmargin=*,itemsep=2pt,topsep=2pt]
  \item \textbf{A pre-deployment calibration pipeline for role-conditioned agents.}
  We combine structured role profiles, contrastive activation extraction,
  coefficient sweeps, and behavioral screening into a practical workflow for
  preparing synthetic agents for social simulation.
  \item \textbf{A large-scale characterization across 275 roles.}
  Across four tested steering strengths and a 228-question role-agnostic
  battery, CastVectors achieve higher judged role-profile
  alignment than an assistant-axis control vector, and most roles show
  increasing alignment with stronger intervention.
  \item \textbf{A failure-aware account of heterogeneous role response.}
  We distinguish controllable, partially deteriorating, and anti-controllable
  roles, and use exploratory geometric-diversity analyses to interpret, rather
  than replace, the behavioral screen.
\end{itemize}

\section{Related Work}
\label{sec:related}

\textbf{LLM social simulation and agent calibration.}
LLM-based simulations now model interactive characters, agent societies,
collective behavior, and cultural or political dynamics at scales that were
previously difficult to instantiate. 
The central methodological question is
not whether language models can produce plausible interaction traces, but
whether those traces reflect the intended population rather than prompt
artifacts, leakage, model bias, or collapsed identity variation. These
concerns make calibration, empirical grounding, interpretability, and
documentation central for LLM-based social simulation.
Existing persona-simulation audits show why these themes matter: prompt
formulations substantially change simulated demographic portrayals
\citep{lutz_prompt_makes_2025}, instruction adherence and personality
measurements are unstable across turns and evaluation settings
\citep{li_measuring_controlling_2024,tosato_persistent_instability_2026}, and
LLM judges need explicit calibration before their role-alignment scores can be
treated as evidence \citep{zheng_judging_llmasajudge_2023,zhou_personaeval_are_2025}.
Our work addresses an upstream calibration problem: we construct, steer, and
screen candidate role-conditioned agents before they are embedded in a larger
simulation.

\textbf{Persona modeling.}
Persona prompting is the most common way to instantiate simulated agents, but
prompt-based control may produce brittle outcomes in practice. \citet{shanahan_roleplay_large_2023} frame
dialogue models as role-playing systems, while \citet{marks_persona_selection_2026}
argue that post-training selects a default Assistant posterior from a broader
latent persona distribution. Empirically, prompt formulations change simulated
demographic portrayals, instruction adherence decays in multi-turn settings,
and personality measurements remain unstable even for large models
\citep{li_measuring_controlling_2024,lutz_prompt_makes_2025,tosato_persistent_instability_2026}. These results motivate role controls that are less dependent on surface prompt wording.\looseness=-1
    
\textbf{Activation steering and persona vectors.}
Activation steering changes behavior by adding a learned direction to model
activations at inference time
\citep{turner_steering_language_2024,panickssery_steering_llama_2024,zou_representation_engineering_2023}. For instance, mean-difference steering has formal support under pointwise-MSE objectives \citep{im_unified_understanding_2025}, but steering reliability depends on the target behavior and coefficient choice \citep{tan_analysing_generalisation_2024}. Persona-vector work shows that internal directions can monitor or control traits \citep{chen_persona_vectors_2025}, scale to large role inventories and identify the Assistant Axis \citep{lu_assistant_axis_2026}, and improve role-specific behavior in smaller role sets \citep{poterti_can_role_2025}. We build on those mechanisms rather than proposing a new steering estimator. In this context, our contribution is a large-scale, failure-aware workflow for constructing and screening candidate role agents.
    
\textbf{Geometry as diagnostics.}
Representation geometry matters here because simulation builders typically
need interpretable failure signals rather than average scores. The linear
representation hypothesis and word-vector arithmetic motivate treating role
directions as structured objects
\citep{park_linear_representation_2024,mikolov_efficient_estimation_2013}. We
use geometry in a secondary role: vector norm, reference distance, RSA, PCA,
and trait projections provide exploratory context for observed response
curves and support, rather than replace, behavioral screening.

\section{A Pre-Deployment Role-Steering Pipeline}
\label{sec:method}

\begin{figure}[t]
    \centering
    \resizebox{\linewidth}{!}{%

    \begin{tikzpicture}[
        pvstep/.style={rectangle, rounded corners=2pt, draw=MLBorderGray, fill=MLLightGray, line width=0.6pt, align=center, text width=1.6cm, minimum height=1.05cm, font=\scriptsize},
        pvpass/.style={rectangle, rounded corners=2pt, draw=MLTeal!70!black, fill=MLTeal!10, line width=0.6pt, align=center, text width=2.0cm, minimum height=0.9cm, font=\scriptsize},
        pvflag/.style={rectangle, rounded corners=2pt, draw=MLAmber!80!black, fill=MLAmber!12, line width=0.6pt, align=center, text width=2.2cm, minimum height=0.9cm, font=\scriptsize},
        pvdeploy/.style={rectangle, rounded corners=2pt, draw=MLTeal!70!black, fill=MLTeal!10, line width=0.6pt, align=center, text width=1.6cm, minimum height=0.9cm, font=\scriptsize},
        pvarrow/.style={->, line width=0.8pt, draw=MLGray}
    ]
      \node[pvstep, fill=MLTeal!10, draw=MLTeal!70!black] (define) at (0,0) {\textbf{Define}\\role profile};
      \node[pvstep, fill=MLBlue!10, draw=MLBlue!70!black] (extract) at (2.2,0) {\textbf{Extract}\\candidate direction};
      \node[pvstep, fill=MLBlue!10, draw=MLBlue!70!black] (sweep) at (4.4,0) {\textbf{Sweep}\\four $\alpha$ values};
      \node[pvstep, fill=MLPurple!10, draw=MLPurple!70!black] (screen) at (6.6,0) {\textbf{Screen}\\alignment and repetition};
      \node[pvpass] (pass) at (9.7,0.65) {\textbf{Pass}\\documented configuration};
      \node[pvflag] (flag) at (9.7,-0.85) {\textbf{Flag}\\revise, lower, prompt, or exclude};
      \node[pvdeploy] (deploy) at (12.15,0.65) {\textbf{Deploy}\\screened agent};
      \draw[pvarrow] (define) -- (extract);
      \draw[pvarrow] (extract) -- (sweep);
      \draw[pvarrow] (sweep) -- (screen);
      \draw[pvarrow] (screen.east) -- node[above, font=\scriptsize, text=MLTeal!80!black] {pass} (pass.west);
      \draw[pvarrow] (screen.east) -- node[below, font=\scriptsize, text=MLAmber!80!black] {flag} (flag.west);
      \draw[pvarrow] (pass) -- (deploy);
    \end{tikzpicture}}
    
    \caption{\textbf{Pre-deployment calibration workflow for role-conditioned agents.}
    The workflow above defines the process of defining a candidate behavior followed by judge-filtered contrastive difference methodology to extract the activations, resulting in the output CastVectors. The CastVectors are evaluated at 4 steering strengths during which the screen retains a configuration or flags the role for low-strength use, profile revision, prompt conditioning, or exclusion.
    }  
    
    \label{fig:method-pipeline}
\end{figure}
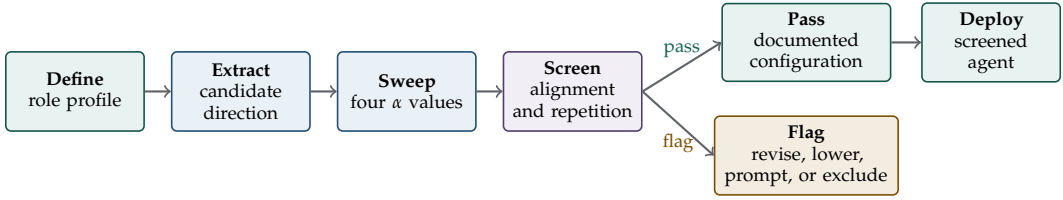

Our pipeline treats every extracted CastVector as a \emph{candidate} role-agent
configuration that must be evaluated before simulation use.
Figure~\ref{fig:method-pipeline} summarizes the operational stages: define a
role profile, extract a candidate vector, sweep the steering coefficient,
screen the resulting behavior, and either pass a documented configuration or
flag it for revision, lower-strength use, prompt conditioning, or exclusion.

\textbf{Model and layer.}
All experiments use OLMo-3-7B-Instruct \citep{team_olmo_olmo_3_2026} on the
residual stream at layer~16, the middlemost layer, following the
middle-layer convention common in role-steering work
\citep{lu_assistant_axis_2026}.\looseness=-1

\textbf{Role inventory and elicitation.}
The inventory contains the 275 role labels of \citet{lu_assistant_axis_2026}. Many entries correspond to
occupations, while others are broader archetypal or nonoccupational roles; we
therefore refer to the collection as a mixed role inventory. For each role we
construct a structured profile of fifteen behavioral directives (e.g.\
\textit{core drive}, \textit{decision style}, \textit{conflict stance},
\textit{risk orientation}, \textit{inner contradiction}; the full list is in
Appendix~\ref{app:role_elicitation}), grounded in a role description, a set of
mandatory tasks, and a set of role contexts. From this profile we generate
five system prompts and $50$ role-specific elicitation questions spanning
seven situation types (e.g.\ resource conflict, ambiguity response, identity
under removal; the full list is in Appendix~\ref{app:role_elicitation}).
Profiles and questions are generated with
\texttt{moonshotai/kimi-k2.5:nitro}, seeded from the O*NET occupational
database where an O*NET occupation matches the role. These generated profiles
are explicit modeling assumptions rather than empirical descriptions of
people in an occupation.
The questions probe decision-making trade-offs implied by the profile rather
than testing domain knowledge or surface-level stylistic mimicry. Generation
details are in Appendix~\ref{app:role_elicitation}.

\textbf{Judge-filtered vector extraction.}
For each role we collect responses under prompted and default (Assistant)
conditions and score each response with an LLM judge
(\texttt{openai/gpt-oss-20b}) into four labels:
\textit{fully role-playing}, \textit{partially role-playing},
\textit{task-responsive}, and \textit{non role-playing}. We retain only
\textit{fully role-playing} positives and \textit{Assistant}-style
negatives, then form the role vector $v_r$ as the layer-16
mean-of-differences over the filtered pairs --- the optimal pointwise-MSE
estimator under the formulation of \citet{im_unified_understanding_2025}.\looseness=-1

\textbf{Steering coefficient sweep.}
Following \citet{lu_assistant_axis_2026}, we steer at the same layer used for
extraction with additive intervention $h \mapsto h + \alpha \, v_r$,
where $h$ is the residual-stream activation at that layer, at
$\alpha \in \{1.0, 1.5, 2.0, 2.5\}$. This grid provides four tested
intervention strengths, so the coefficient-response claims in this paper are
limited to these values. The reported coefficient-response screen starts at
$\alpha=1.0$; separate prompted-reference or unsteered diagnostics are not part
of the four-point correlation used to classify controllability. Decoding
settings and intervention-span details are documented with the released
artifacts.

\textbf{Role-profile alignment readout and screening.}
For every (role, $\alpha$) cell we generate responses to $228$ role-agnostic
alignment questions and score them with GPT-4.1-mini along six dimensions:
an overall role-profile alignment score plus emotional register, vocab choice,
social dynamic, motivation, and worldview alignment against a prompted role
reference. GPT-4.1-mini generates the prompted references and serves as all
evaluation judges, so shared model priors may influence both the target and
the score.
The screen reports three quantities that matter for simulation use: absolute
role-profile alignment, response across the four tested coefficients, and text
quality indicators such as unique-bigram ratio as a lexical repetition proxy.
We also compare every role to an assistant-axis directional control: the
vector produced for that role by the extraction pipeline of
\citet{lu_assistant_axis_2026}, which is near-parallel across roles
(Appendix~\ref{app:controllability}), under the same questions and
coefficients. The
assistant-axis comparison is not scale matched because its vectors have larger
mean $\ell_2$ norm than the role-specific vectors (9.68 versus 3.79). Full
judge prompts, pairwise position-swap checks, and reliability analyses are in
Appendix~\ref{app:eval_protocols}; the behavioral RDM split-half result
($r{=}0.97$ for correlation distance) measures internal stability of this
automated readout, not external validity.

\section{Evaluation Across 275 Roles}
\label{sec:steering-at-scale}
We evaluate the pipeline with three questions relevant to simulation builders:
how judged role-profile alignment changes across the tested coefficient grid,
how CastVectors compare with the assistant-axis
control vectors with respect to role alignment, and which candidate roles deteriorate under stronger intervention.
Full per-metric tables and per-$\alpha$ breakdowns are in
Appendix~\ref{app:judge_results}; controllability statistics are in
Appendix~\ref{app:controllability}.

\begin{figure}[t]
  \centering
  \begin{minipage}{0.495\linewidth}
    \centering
    \includegraphics[width=\linewidth]{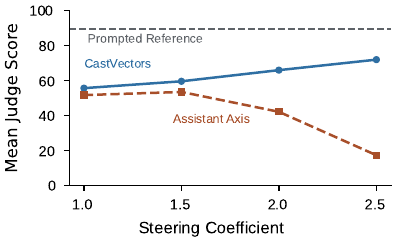}
  \end{minipage}\hfill
  \begin{minipage}{0.495\linewidth}
    \centering
    \includegraphics[width=\linewidth]{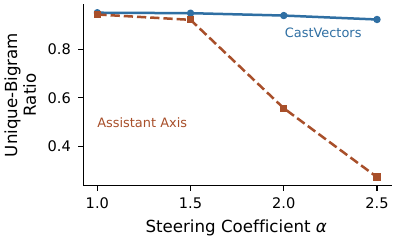}
  \end{minipage}
  
  \caption{Aggregate evaluation across $n{=}275$ roles. \textbf{Left}: mean judged role-profile alignment across all roles and across $\alpha \in \{1.0,1.5,2.0,2.5\}$. CastVectors perform better across steering coefficients, with the gap increasing at every interval. \textbf{Right}: unique bigram ratio is used as a lexical repetition proxy. CastVectors display less repetition and thus lower inclination for incohesiveness across all responses for a role given a certain $\alpha$. Per-metric score tables are in Appendix~\ref{app:judge_results}.}
  \label{fig:headline-empirical}
\end{figure}

\textbf{CastVectors show higher role-profile alignment than the assistant-axis control.}
Aggregated across $275$ roles and $228$ questions per (role, $\alpha$) cell,
the role-specific condition receives a mean overall role-profile alignment
score of $63.2$, compared with $41.1$ for the assistant-axis directional
control and $89.2$ for the prompted role reference. The reference is a
model-generated prompting condition rather than empirical ground truth. The
assistant-axis result should be read with a scale caveat: its vectors have a
larger mean $\ell_2$ norm than the CastVectors (9.68 versus 3.79),
so equal coefficients do not produce equal perturbation magnitudes. Under the current extraction and
evaluation setup, the assistant-related control does not provide a pure directional comparison, but rather a behavioral and role-aligned comparison of the methodologies as applied. It should not
be read as evidence that assistant-like directions point away from natural
assistant behavior; the control also has a larger norm and degrades most at the
largest coefficients. Unique-bigram
ratio remains high for the CastVectors over the tested range and
falls sharply for the assistant-axis control vector at the largest
coefficients; this statistic is a lexical repetition proxy, not a complete
measure of semantic quality (\reffig{fig:headline-empirical}, right).

\begin{table}[t]
\centering
\renewcommand{\arraystretch}{1.3}
\setlength{\tabcolsep}{6pt}
\small
\begin{tabular}{l c}
\toprule
\textbf{Condition} & \textbf{Overall score (mean)} \\
\midrule
Prompted role reference & $89.2$ \\
Role-specific direction, selected best mean ($\alpha=2.5$) & $71.9$ \\
Assistant-axis directional control, selected best mean ($\alpha=1.5$) & $53.4$ \\
\bottomrule
\end{tabular}
\caption{Selected operating points for the prompted role reference,
role-specific condition, and assistant-axis directional control. The
role-specific and assistant-axis rows each use the highest mean overall score
within that condition's tested grid; this is not a matched-coefficient or
scale-matched comparison.}
\label{tab:baseline_comparison}
\end{table}

\begin{figure}[t]
  \centering
  \begin{minipage}[t]{0.495\linewidth}
    \centering
    \includegraphics[width=\linewidth]{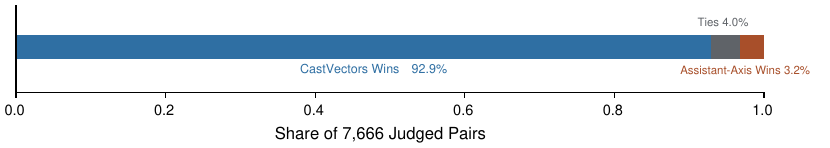}
  \end{minipage}\hfill
  \begin{minipage}[t]{0.495\linewidth}
    \centering
    \includegraphics[width=\linewidth]{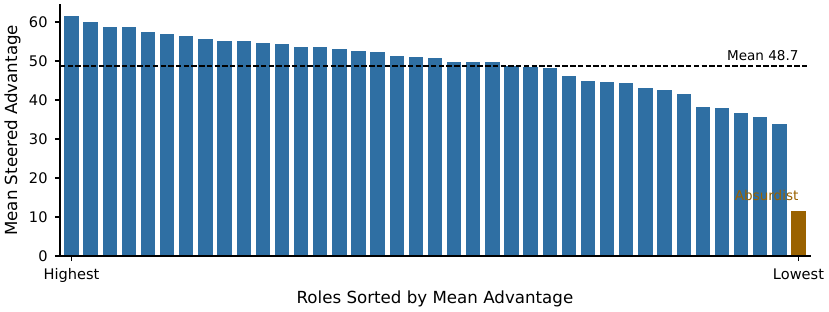}
  \end{minipage}
  \caption{Pairwise preference check for the 39-role subset at
  $\alpha{=}2.5$. \textbf{Left}: role-specific steered responses win 92.9\% of 7,666
  comparisons; ties account for 4.0\% and assistant-axis wins for 3.2\%.
  \textbf{Right}: the debiased steered advantage is positive for all 39 evaluated roles
  (mean of per-role means 48.7 points; pooled mean 47.9; minimum 11.4). The
  comparison uses the same non-scale-matched assistant-axis directional control
  as Figure~\ref{fig:headline-empirical}.}
  \label{fig:pairwise-summary}
\end{figure}

Pairwise checks on the 39-role subset give the same practical conclusion at
$\alpha{=}2.5$. Across 7,666 response pairs, role-specific steered responses
win 92.9\% of comparisons against the assistant-axis directional control, and
the debiased advantage is positive for all 39 evaluated roles
(Figure~\ref{fig:pairwise-summary}). This is still a comparison to the
control vector rather than a prompt-only or norm-matched baseline, but it
shows that the aggregate score gap is also visible to the pairwise judge.\looseness=-1

\FloatBarrier

\textbf{CastVectors should be screened for alignment.}
Aggregate means are not enough for simulation deployment. At every $\alpha$,
some roles receive high role-profile alignment scores, while others stagnate
or degrade. Screening only at a single coefficient would miss this distinction:
low-strength alignment can coexist with high-strength deterioration, while a
high-strength-only report can hide usable lower-coefficient configurations. A
role inventory therefore needs a per-role screen: a vector is useful only if it
expresses the intended profile under the tested questions without increasing
lexical repetition or pushing the model away from the role at stronger
coefficients.\looseness=-1

\subsection{Most roles improve across the tested coefficient grid}
\label{sec:alpha-response}
\label{sec:controllability}

\begin{figure}[h]
  \centering
  \includegraphics[width=\linewidth]{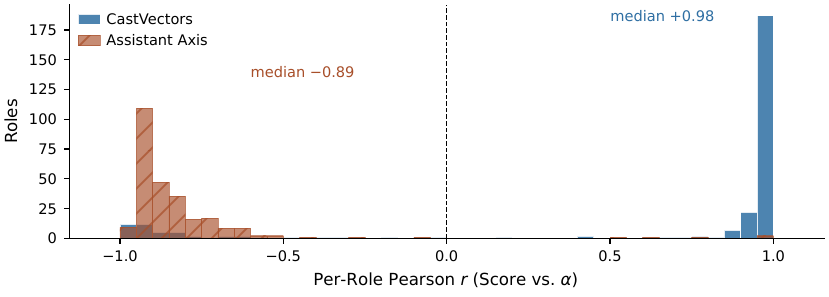}
  \caption{Per-role Pearson $r$ between steering coefficient $\alpha$ and
    overall role-profile alignment score across four tested coefficients.
    Role-specific directions have median $r{=}{+}0.98$, and 74\% of roles
    improve at every consecutive step; the assistant-axis directional control
    from \citet{lu_assistant_axis_2026} has median $r{=}{-}0.89$. Because each
    correlation is based on four values and the two vector families are not
    norm matched, the plot characterizes within-grid response rather than
    behavior beyond the evaluated coefficients or a scale-controlled causal
    comparison.}
  \label{fig:pearson_r_dist}
\end{figure}

For a simulation user, stronger intervention should usually produce stronger
role-profile expression rather than generic drift. We measure this per role by
the Pearson correlation between $\alpha$ and overall role-profile alignment
score across the four tested coefficients.

The distribution is strongly bimodal (Figure~\ref{fig:pearson_r_dist}). Most
roles behave as desired: median $r{=}+0.98$, $82\%$ are positive, $79\%$
exceed $r{=}0.8$, and $74\%$ improve monotonically across all four $\alpha$
steps. These statistics show that $\alpha$ is a useful within-grid tuning
parameter for the majority of the inventory. The four-point correlation should
not be extrapolated beyond the evaluated coefficient grid. The minority mode
near $r{=}-1$ is just as important: it exposes roles that should be handled
specially rather than silently mixed into a simulated population.

\begin{center}
\centering
\small
\begin{tabular}{lcc}
\toprule
\textbf{Category} & \textbf{\% of 275} & \textbf{Count} \\
\midrule
Controllable (overall $r \geq 0$)                        & 82\% & 225 \\
Partial deterioration (overall $r < 0$, not all six)     & 4\%  & 12  \\
Anti-controllable over tested range (all six axes $r < 0$) & 14\% & 38  \\
\bottomrule
\end{tabular}
\captionof{table}{Operational response categories across the 275-role inventory.
Categories are defined by score response over
$\alpha \in \{1.0,1.5,2.0,2.5\}$ and do not identify the mechanism of
deterioration. The 4\% partial-deterioration row is the population whose
aggregate score declines under steering while at least one sub-dimension still
responds positively to $\alpha$.}
\label{tab:controllability-categories}
\end{center}

\subsection{Screening identifies roles that deteriorate under stronger steering}
\label{sec:anti-controllable}
\label{sec:anticontrollability}

\begin{figure}[h]
  \centering
  \begin{minipage}{0.495\linewidth}
    \centering
    \includegraphics[width=\linewidth]{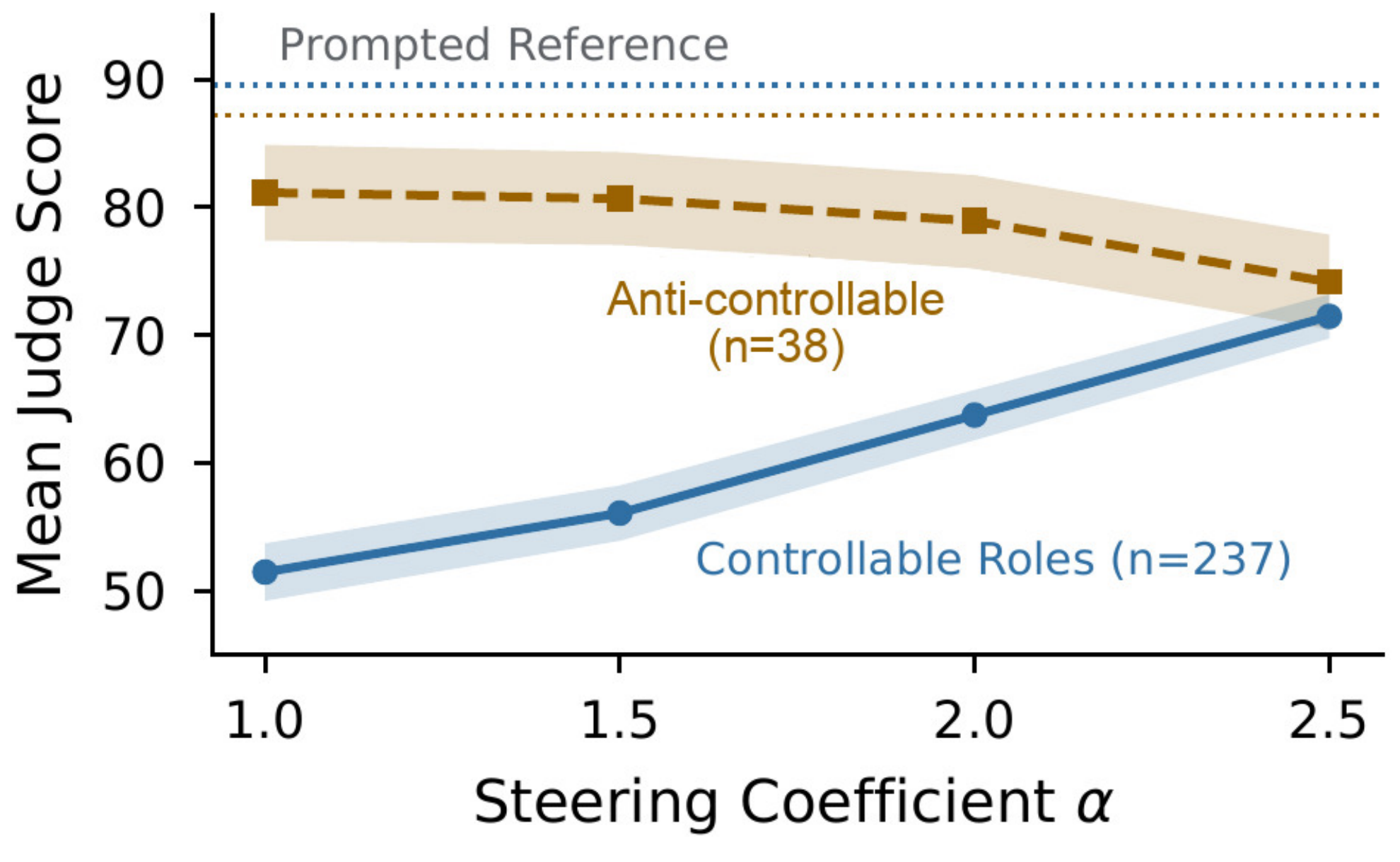}
  \end{minipage}\hfill
  \begin{minipage}{0.495\linewidth}
    \centering
    \includegraphics[width=\linewidth]{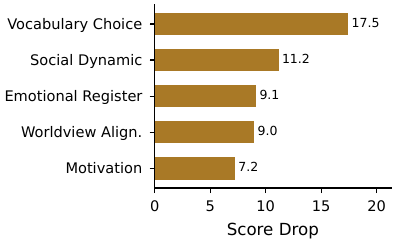}
  \end{minipage}
  \caption{Response of roles classified as anti-controllable over the tested
    range. \textbf{Left:} mean judge score vs.\ $\alpha$ for anti-controllable
    (orange) and remaining (blue) roles, with each group's prompted reference
    level (dotted); shading is the 95\% interval of the group mean. The 38
    anti-controllable roles decline monotonically to $74.2$, while the
    remaining roles rise. \textbf{Right:} mean score drop per measured
    dimension from $\alpha{=}1.0$ to $2.5$. 
    The figure
    documents deterioration under stronger intervention, not failure at the
    lowest tested coefficient; early saturation, over-steering,
    target-direction mismatch, and other mechanisms remain possible
    interpretations.} 
  \label{fig:anticontrollability}
\end{figure}

The coefficient sweep identifies $38$ roles ($14\%$ of 275) whose score
decreases with $\alpha$ across all six measured axes. We call these roles
\emph{anti-controllable over the tested range}. They include analyst, coach,
consultant, moderator, supervisor, teacher, and validator, but the full
inventory also contains nonoccupational and archetypal entries; the category
is an operational role-vector result rather than a finding about professions
as real populations. The pattern is structured across roles and dimensions
under the current readout. It is an operational warning that stronger
intervention is not uniformly beneficial.

At $\alpha{=}1.0$, these anti-controllable roles already average $81.1$ judged
role-profile alignment, versus $55.6$ across all roles and within $6.0$ points
of their own prompted references. With stronger steering, this average declines
to $74.2$ at $\alpha{=}2.5$ (Figure~\ref{fig:anticontrollability}; per-role and per-dimension breakdowns in
Appendix~\ref{app:anticontrollability}). This pattern is
consistent with poor vector alignment or over-steering, but the present four-point
screen does not distinguish those mechanisms: all four points are steered
conditions, and the prompted-reference baseline is a separate model-generated
target rather than a comparable point on the same coefficient trajectory. The
practical rule is independent of the mechanism: candidate role directions
should be swept and screened before agents are used in a simulation.\looseness=-1

\section{Geometry as Diagnostic Context}
\label{sec:diagnostics}
\label{sec:geometry}
The behavioral sweep is the primary screen. We use representation geometry as
a diversity and failure-context diagnostic: it asks whether the role-vector
inventory preserves separations relevant to heterogeneous simulation, and how
those directions relate to the observed coefficient-response curves. Across
roles, vector norm is moderately
associated with the fitted response slope ($r{=}+0.49$). Distance to the pipeline-extracted Assistant role vector is also inversely correlated with the score difference to the reference, with the strongest correlation at
$\alpha{=}1.0$ ($r{=}-0.419$) and a weaker correlation at $\alpha{=}2.5$
($r{=}-0.145$). This relation still holds after accounting for vector magnitude, indicating that the association is directional rather than driven by magnitude alone.

Pairwise representational similarity has a modest association with pairwise
behavioral similarity. The cosine/correlation RSA rises from $\rho{=}+0.085$
at $\alpha{=}1.0$ to $\rho{=}+0.179$ at $\alpha{=}2.5$ under the current
readout. PCA, trait-axis projections, and rank analyses in the appendix
provide additional descriptions of the role-vector cloud, but their effect
sizes are moderate and their semantic labels are interpretive.

These analyses support the behavioral screen by adding post hoc context for
why some directions may require different coefficient choices and by making
the assumptions behind the role inventory easier to inspect. They are
diagnostics rather than deployment rules. The full technical account is in
Appendix~\ref{sec:geometry-behavior}.

\section{Implications, Limitations, and Responsible Use}
\label{sec:discussion}

\paragraph{What the screen establishes.}
The study demonstrates a large-scale procedure for constructing and screening
role-conditioned agents under one model, one layer, one coefficient grid, and
one evaluation pipeline. CastVectors receive higher mean judged
role-profile alignment than the assistant-axis directional control across the
tested coefficients, and most roles improve as intervention strength grows.
The role-level analysis is equally important: 38 roles decline across every
measured dimension, showing that a single global steering strength is not
appropriate for the full inventory. The screen asks which coefficient, if any,
is usable for each role.

\paragraph{For simulation builders.}
The workflow is meant to sit before the simulation run itself. A builder
defines the intended synthetic role and documents the profile assumptions;
extracts a candidate activation direction; evaluates the candidate over the
tested coefficient grid; inspects role-profile alignment and lexical
repetition proxies; flags declining or dimensionally inconsistent roles; and
records the profile, coefficient, reference condition, and screening result
before using the agent in a simulation. The output is a documented agent
configuration and a set of role-level warnings.

\paragraph{Validity boundaries.}
Role-profile alignment is an operational screen, not a population-validity
study. The evaluation measures agreement with constructed role profiles and
prompted references rather than correspondence to real workers or real
occupational populations. GPT-4.1-mini generated the references and served as
the evaluator, so shared priors may influence the target and the score. The
assistant-axis comparison is directional rather than scale matched, the
controllability classification uses four nonzero values, and all experiments use
OLMo-3-7B-Instruct at layer 16. The current study leaves prompt paraphrase
robustness, multi-turn persistence, cross-model generality, agent-agent
interaction, and downstream simulation fidelity for future work. The geometry
analyses are correlational diagnostic context rather than prospective
selection rules.

\paragraph{Responsible interpretation.}
The role profiles and CastVectors are model- and prompt-derived archetypes. They
are not psychological representations of people in an occupation. Real roles
contain substantial variation across individuals, institutions, cultures,
seniority levels, and contexts, while one vector per label can flatten that
variation or amplify familiar stereotypes. Simulation reports should therefore
disclose how roles were defined, which coefficient was selected, which agents
were flagged, and which outcomes depend on these modeling choices.\looseness=-1

\section{Conclusion}
\label{sec:conclusion}

We present activation steering as a screening workflow for constructing
role-conditioned synthetic agents before simulation use. Across a 275-role
inventory, role-specific directions achieve higher judged role-profile
alignment than an assistant-axis directional control over four tested steering
strengths, and most roles improve as the coefficient grows. A notable
minority declines across every measured dimension, demonstrating that
candidate directions should be swept and screened rather than deployed with a
uniform setting. The contribution is methodological: activation steering
produces candidate configurations that can be measured, compared, documented,
and rejected before they are used in a simulation. For social-simulation
builders, the practical lesson is to treat role construction as a screened
configuration step rather than a one-shot prompt or vector choice.

\ifcolmsubmission
\else
\section*{Acknowledgments}

This research was supported in part through research cyberinfrastructure
resources and services provided by the Partnership for an Advanced Computing
Environment (PACE) at the Georgia Institute of Technology, Atlanta, Georgia,
USA. RRID:SCR\_027619. We additionally thank the PACE team for the operational
support that enabled the role-steering experiments in this work.

This work was done as part of the
\href{https://www.matsprogram.org/}{ML Alignment \& Theory Scholars (MATS)
9.0 Fellowship}.

We thank \href{https://modal.com/}{Modal Labs} and CEO Charles Frye for
additional compute support and for providing infrastructure that accelerated
the steering and evaluation experiments.

\fi

\bibliographystyle{colm2026_conference}
\bibliography{references}

\begin{thebibliography}{20}
\providecommand{\natexlab}[1]{#1}
\providecommand{\url}[1]{\texttt{#1}}
\expandafter\ifx\csname urlstyle\endcsname\relax
  \providecommand{\doi}[1]{doi: #1}\else
  \providecommand{\doi}{doi: \begingroup \urlstyle{rm}\Url}\fi

\bibitem[Bas \& Novak(2026)Bas and Novak]{bas_what_can_2026}
Tetiana Bas and Krystian Novak.
\newblock What {{Can We Actually Steer}}? {{A Multi-Behavior Study}} of
  {{Activation Control}}, January 2026.

\bibitem[Chen et~al.(2025)Chen, Arditi, Sleight, Evans, and
  Lindsey]{chen_persona_vectors_2025}
Runjin Chen, Andy Arditi, Henry Sleight, Owain Evans, and Jack Lindsey.
\newblock Persona {{Vectors}}: {{Monitoring}} and {{Controlling Character
  Traits}} in {{Language Models}}, September 2025.

\bibitem[Im \& Li(2025)Im and Li]{im_unified_understanding_2025}
Shawn Im and Sharon Li.
\newblock A {{Unified Understanding}} and {{Evaluation}} of {{Steering
  Methods}}, February 2025.

\bibitem[Kim et~al.(2024)Kim, Shin, Cho, Jang, Longpre, Lee, Yun, Shin, Kim,
  Thorne, and Seo]{kim_prometheus_inducing_2023}
Seungone Kim, Jamin Shin, Yejin Cho, Joel Jang, Shayne Longpre, Hwaran Lee,
  Sangdoo Yun, Seongjin Shin, Sungdong Kim, James Thorne, and Minjoon Seo.
\newblock Prometheus: {{Inducing Fine-Grained Evaluation Capability}} in
  {{Language Models}}.
\newblock In \emph{The {{Twelfth International Conference}} on {{Learning
  Representations}}}, 2024.
\newblock \doi{10.48550/arXiv.2310.08491}.

\bibitem[Li et~al.(2024)Li, Liu, Bashkansky, Bau, Vi{\'e}gas, Pfister, and
  Wattenberg]{li_measuring_controlling_2024}
Kenneth Li, Tianle Liu, Naomi Bashkansky, David Bau, Fernanda Vi{\'e}gas,
  Hanspeter Pfister, and Martin Wattenberg.
\newblock Measuring and {{Controlling Instruction}} ({{In}}){{Stability}} in
  {{Language Model Dialogs}}.
\newblock In \emph{First {{Conference}} on {{Language Modeling}}}, July 2024.
\newblock \doi{10.48550/arXiv.2402.10962}.

\bibitem[Lu et~al.(2026)Lu, Gallagher, Michala, Fish, and
  Lindsey]{lu_assistant_axis_2026}
Christina Lu, Jack Gallagher, Jonathan Michala, Kyle Fish, and Jack Lindsey.
\newblock The {{Assistant Axis}}: {{Situating}} and {{Stabilizing}} the
  {{Default Persona}} of {{Language Models}}, January 2026.

\bibitem[Lutz et~al.(2025)Lutz, Sen, Ahnert, Rogers, and
  Strohmaier]{lutz_prompt_makes_2025}
Marlene Lutz, Indira Sen, Georg Ahnert, Elisa Rogers, and Markus Strohmaier.
\newblock The {{Prompt Makes}} the {{Person}}(a): {{A Systematic Evaluation}}
  of {{Sociodemographic Persona Prompting}} for {{Large Language Models}}.
\newblock In \emph{Findings of the {{Association}} for {{Computational
  Linguistics}}: {{EMNLP}} 2025}, pp.\  23212--23237, Suzhou, China, October
  2025. Association for Computational Linguistics.
\newblock \doi{10.18653/v1/2025.findings-emnlp.1261}.

\bibitem[Marks et~al.(2026)Marks, Lindsey, and
  Olah]{marks_persona_selection_2026}
Sam Marks, Jack Lindsey, and Christopher Olah.
\newblock The {{Persona Selection Model}}: {{Why AI Assistants}} might
  {{Behave}} like {{Humans}}.
\newblock https://alignment.anthropic.com/2026/psm/, February 2026.

\bibitem[Mikolov et~al.(2013)Mikolov, Chen, Corrado, and
  Dean]{mikolov_efficient_estimation_2013}
Tomas Mikolov, Kai Chen, Greg Corrado, and Jeffrey Dean.
\newblock Efficient {{Estimation}} of {{Word Representations}} in {{Vector
  Space}}, September 2013.

\bibitem[Panickssery et~al.(2024)Panickssery, Gabrieli, Schulz, Tong, Hubinger,
  and Turner]{panickssery_steering_llama_2024}
Nina Panickssery, Nick Gabrieli, Julian Schulz, Meg Tong, Evan Hubinger, and
  Alexander~Matt Turner.
\newblock Steering {{Llama}} 2 via {{Contrastive Activation Addition}}.
\newblock In \emph{Proceedings of the 62nd {{Annual Meeting}} of the
  {{Association}} for {{Computational Linguistics}} ({{Volume}} 1: {{Long
  Papers}})}, pp.\  15504--15522, Bangkok, Thailand, July 2024. Association for
  Computational Linguistics.
\newblock \doi{10.18653/v1/2024.acl-long.828}.

\bibitem[Park et~al.(2024)Park, Choe, and
  Veitch]{park_linear_representation_2024}
Kiho Park, Yo~Joong Choe, and Victor Veitch.
\newblock The {{Linear Representation Hypothesis}} and the {{Geometry}} of
  {{Large Language Models}}.
\newblock In \emph{Proceedings of the 41st {{International Conference}} on
  {{Machine Learning}}}, volume 235 of \emph{Proceedings of {{Machine Learning
  Research}}}, pp.\  39643--39666. PMLR, July 2024.
\newblock \doi{10.48550/arXiv.2311.03658}.

\bibitem[Potert{\`i} et~al.(2025)Potert{\`i}, Seveso, and
  Mercorio]{poterti_can_role_2025}
Daniele Potert{\`i}, Andrea Seveso, and Fabio Mercorio.
\newblock Can {{Role Vectors Affect LLM Behaviour}}?
\newblock In Christos Christodoulopoulos, Tanmoy Chakraborty, Carolyn Rose, and
  Violet Peng (eds.), \emph{Findings of the {{Association}} for {{Computational
  Linguistics}}: {{EMNLP}} 2025}, pp.\  17735--17747, Suzhou, China, November
  2025. Association for Computational Linguistics.
\newblock ISBN 979-8-89176-335-7.
\newblock \doi{10.18653/v1/2025.findings-emnlp.963}.

\bibitem[Shanahan et~al.(2023)Shanahan, McDonell, and
  Reynolds]{shanahan_roleplay_large_2023}
Murray Shanahan, Kyle McDonell, and Laria Reynolds.
\newblock Role-{{Play}} with {{Large Language Models}}.
\newblock \emph{Nature}, 623\penalty0 (7987):\penalty0 493--498, November 2023.
\newblock \doi{10.1038/s41586-023-06647-8}.

\bibitem[Tan et~al.(2024)Tan, Chanin, Lynch, Paige, Kanoulas, {Garriga-Alonso},
  and Kirk]{tan_analysing_generalisation_2024}
Daniel Tan, David Chanin, Aengus Lynch, Brooks Paige, Dimitrios Kanoulas,
  Adri{\`a} {Garriga-Alonso}, and Robert Kirk.
\newblock Analysing the {{Generalisation}} and {{Reliability}} of {{Steering
  Vectors}}.
\newblock In \emph{Proceedings of the 38th {{International Conference}} on
  {{Neural Information Processing Systems}}}, volume~37 of \emph{{{NIPS}} '24},
  pp.\  139179--139212, Red Hook, NY, USA, December 2024. Curran Associates
  Inc.
\newblock ISBN 979-8-3313-1438-5.
\newblock \doi{10.52202/079017-4417}.

\bibitem[{Team Olmo} et~al.(2026){Team Olmo}, Ettinger, Bertsch, Kuehl, Graham,
  Heineman, Groeneveld, Brahman, Timbers, Ivison, Morrison, Poznanski, Lo,
  Soldaini, Jordan, Chen, Noukhovitch, Lambert, Walsh, Dasigi, Berry, Malik,
  Shah, Geng, Arora, Gupta, Anderson, Xiao, Murray, Romero, Graf, Asai, Bhagia,
  Wettig, Liu, Rangapur, Anastasiades, Huang, Schwenk, Trivedi, Magnusson,
  Lochner, Liu, Miranda, Sap, Morgan, Schmitz, Guerquin, Wilson, Huff, Le~Bras,
  Xin, Shao, Skjonsberg, Shen, Li, Wilde, Pyatkin, Merrill, Chang, Gu, Zeng,
  Sabharwal, Zettlemoyer, Koh, Farhadi, Smith, and
  Hajishirzi]{team_olmo_olmo_3_2026}
{Team Olmo}, Allyson Ettinger, Amanda Bertsch, Bailey Kuehl, David Graham,
  David Heineman, Dirk Groeneveld, Faeze Brahman, Finbarr Timbers, Hamish
  Ivison, Jacob Morrison, Jake Poznanski, Kyle Lo, Luca Soldaini, Matt Jordan,
  Mayee Chen, Michael Noukhovitch, Nathan Lambert, Pete Walsh, Pradeep Dasigi,
  Robert Berry, Saumya Malik, Saurabh Shah, Scott Geng, Shane Arora, Shashank
  Gupta, Taira Anderson, Teng Xiao, Tyler Murray, Tyler Romero, Victoria Graf,
  Akari Asai, Akshita Bhagia, Alexander Wettig, Alisa Liu, Aman Rangapur, Chloe
  Anastasiades, Costa Huang, Dustin Schwenk, Harsh Trivedi, Ian Magnusson,
  Jaron Lochner, Jiacheng Liu, Lester James~V. Miranda, Maarten Sap, Malia
  Morgan, Michael Schmitz, Michal Guerquin, Michael Wilson, Regan Huff, Ronan
  Le~Bras, Rui Xin, Rulin Shao, Sam Skjonsberg, Shannon~Zejiang Shen,
  Shuyue~Stella Li, Tucker Wilde, Valentina Pyatkin, Will Merrill, Yapei Chang,
  Yuling Gu, Zhiyuan Zeng, Ashish Sabharwal, Luke Zettlemoyer, Pang~Wei Koh,
  Ali Farhadi, Noah~A. Smith, and Hannaneh Hajishirzi.
\newblock Olmo 3, April 2026.

\bibitem[Tosato et~al.(2026)Tosato, Helbling, {Mantilla-Ramos}, Hegazy, Tosato,
  Lemay, Rish, and Dumas]{tosato_persistent_instability_2026}
Tommaso Tosato, Saskia Helbling, Yorguin-Jose {Mantilla-Ramos}, Mahmood Hegazy,
  Alberto Tosato, David~John Lemay, Irina Rish, and Guillaume Dumas.
\newblock Persistent {{Instability}} in {{LLM}}'s {{Personality Measurements}}:
  {{Effects}} of {{Scale}}, {{Reasoning}}, and {{Conversation History}}.
\newblock In \emph{Proceedings of the {{AAAI Conference}} on {{Artificial
  Intelligence}}}, volume~40, pp.\  37961--37969. AAAI Press, March 2026.
\newblock \doi{10.1609/aaai.v40i44.41133}.

\bibitem[Turner et~al.(2024)Turner, Thiergart, Leech, Udell, Vazquez, Mini, and
  MacDiarmid]{turner_steering_language_2024}
Alexander~Matt Turner, Lisa Thiergart, Gavin Leech, David Udell, Juan~J.
  Vazquez, Ulisse Mini, and Monte MacDiarmid.
\newblock Steering {{Language Models With Activation Engineering}}, October
  2024.

\bibitem[Zheng et~al.(2023)Zheng, Chiang, Sheng, Zhuang, Wu, Zhuang, Lin, Li,
  Li, Xing, Zhang, Gonzalez, and Stoica]{zheng_judging_llmasajudge_2023}
Lianmin Zheng, Wei-Lin Chiang, Ying Sheng, Siyuan Zhuang, Zhanghao Wu, Yonghao
  Zhuang, Zi~Lin, Zhuohan Li, Dacheng Li, Eric~P. Xing, Hao Zhang, Joseph~E.
  Gonzalez, and Ion Stoica.
\newblock Judging {{LLM-as-a-Judge}} with {{MT-Bench}} and {{Chatbot Arena}}.
\newblock In \emph{Advances in {{Neural Information Processing Systems}} 36},
  2023.
\newblock \doi{10.48550/arXiv.2306.05685}.

\bibitem[Zhou et~al.(2025)Zhou, Zhang, Gao, Jiang, and
  Wang]{zhou_personaeval_are_2025}
Lingfeng Zhou, Jialing Zhang, Jin Gao, Mohan Jiang, and Dequan Wang.
\newblock {{PersonaEval}}: {{Are LLM Evaluators Human Enough}} to {{Judge
  Role-Play}}?, August 2025.

\bibitem[Zou et~al.(2023)Zou, Phan, Chen, Campbell, Guo, Ren, Pan, Yin,
  Mazeika, Dombrowski, Goel, Li, Byun, Wang, Mallen, Basart, Koyejo, Song,
  Fredrikson, Kolter, and Hendrycks]{zou_representation_engineering_2023}
Andy Zou, Long Phan, Sarah Chen, James Campbell, Phillip Guo, Richard Ren,
  Alexander Pan, Xuwang Yin, Mantas Mazeika, Ann-Kathrin Dombrowski, Shashwat
  Goel, Nathaniel Li, Michael~J. Byun, Zifan Wang, Alex Mallen, Steven Basart,
  Sanmi Koyejo, Dawn Song, Matt Fredrikson, J.~Zico Kolter, and Dan Hendrycks.
\newblock Representation {{Engineering}}: {{A Top-Down Approach}} to {{AI
  Transparency}}, October 2023.

\end{thebibliography}

\newpage
\appendix

\section{Datasets and Prompts}
\label{app:datasets_prompts}

\textbf{Takeaway.} The role-steering recipe depends on a fixed mixed role
inventory, role-specific elicitation prompts, and role-agnostic evaluation
questions. This appendix records those sources and prompts so the main-text
role-profile alignment screen is reproducible.

\subsection{Role Profiles and Elicitation Battery}
\label{app:role_elicitation}

Role profiles and elicitation questions were generated with
\texttt{moonshotai/kimi-k2.5:nitro}; the generating model is recorded
per file in the released role-dataset artifacts. Each role is seeded
with an occupation description and task list --- drawn from the O*NET
occupational database where an O*NET occupation matches the role ---
from which a structured psychological profile of fifteen behavioral
directives is generated: \textit{core drive}, \textit{decision style},
\textit{non-negotiable}, \textit{conflict stance}, \textit{social
posture}, \textit{recurring resentment}, \textit{risk orientation},
\textit{failure response}, \textit{instinctive blame target},
\textit{value hierarchy}, \textit{cognitive bias}, \textit{rejected
premise}, \textit{inner contradiction}, \textit{attention pattern},
and \textit{relationship to authority}. Five system prompts
are then built from disjoint subsets of the profile's directives, and
$50$ role-specific elicitation questions are generated across seven
situation types: \textit{resource conflict}, \textit{ambiguity
response}, \textit{social friction}, \textit{constraint reaction},
\textit{identity under removal}, \textit{unconstrained choice}, and
\textit{competing pulls}.

\subsection{Alignment Battery (Behavioral Readout)}
\label{app:elicitation_battery}

All role evaluations are conducted using a fixed battery of 228 unique open-ended
questions designed to elicit opinion, reasoning, and behavioral expression rather than
factual recall. (The source battery file contains 229 entries; one ethics question is
an exact duplicate and is deduplicated before evaluation.)  The questions are deliberately role-agnostic --- no question names or implies
a specific role --- so that any role-consistent signal in the response is attributable
to the steering intervention rather than the prompt.  The battery spans 14 semantic
categories to ensure broad coverage across distinct registers of expression.

\begin{table}[h]
\centering
\renewcommand{\arraystretch}{1.3}
\setlength{\tabcolsep}{8pt}
\small
\begin{tabular}{lc}
\toprule
\textbf{Category} & \textbf{Questions} \\
\midrule
Social \& political opinion             & 20 \\
Philosophical \& existential            & 20 \\
Life advice \& self-improvement         & 20 \\
Business \& professional strategy       & 20 \\
Personal \& emotional introspective     & 20 \\
Ethics \& moral philosophy              & 17 \\
Interpersonal conflict \& relationships & 19 \\
Judgment of others' behavior            & 9  \\
Financial decision-making               & 8  \\
Occupational epistemics                 & 15 \\
Failure attribution                     & 15 \\
Time, risk, \& novelty tradeoffs        & 15 \\
Authority \& hierarchy                  & 15 \\
Social energy \& interaction            & 15 \\
\midrule
\textbf{Total}                          & \textbf{228} \\
\bottomrule
\end{tabular}
\caption{Alignment battery composition across 14 semantic categories, after
removing one duplicated ethics question from the 229-entry source file.}
\label{tab:battery_categories}
\end{table}

\paragraph{Representative questions by category.}

\begin{description}[leftmargin=0pt, itemsep=4pt]

\item[\textit{Social \& political opinion}]
``How should society prepare for technological disruption?'' /
``Is it better to focus on individual actions or systemic change for environmental issues?'' /
``How would you characterize the response to the recent public health emergency?''

\item[\textit{Philosophical \& existential}]
``What's your view on the role of fate versus free will in human life?'' /
``What is the relationship between language and thought?'' /
``What is the concept of happiness?''

\item[\textit{Life advice \& self-improvement}]
``I'm feeling overwhelmed with my responsibilities.'' /
``I'm scared to leave my comfort zone even though I'm unhappy where I am.'' /
``What's the most important thing to focus on in your twenties?''

\item[\textit{Business \& professional strategy}]
``A senior partner at your firm questions your client recommendations. What's your approach?'' /
``How do different leadership styles impact workplace productivity?'' /
``What makes some renewable energy projects successful while others fail?''

\item[\textit{Personal \& emotional introspective}]
``What makes you feel most insecure about yourself?'' /
``Describe what it's like when you finally understand something complex.'' /
``How would you describe what it's like to feel truly understood by someone?''

\item[\textit{Occupational epistemics}]
``When someone presents you with data that contradicts your professional experience,
how do you decide which to trust?'' /
``How do you distinguish between something that works and something you can prove works?'' /
``When two credible experts disagree, how do you form your own position?''

\item[\textit{Failure attribution}]
``When a project you led doesn't achieve its goals, what's your first instinct about
where things went wrong?'' /
``How do you distinguish between a failure caused by bad luck and one caused by poor judgment?'' /
``Describe how you approach a post-mortem or review after something goes wrong.''

\item[\textit{Time, risk, \& novelty tradeoffs}]
``When facing a tight deadline, what do you sacrifice first: thoroughness, communication, or sleep?'' /
``At what point does the risk of inaction outweigh the risk of making a potentially wrong decision?'' /
``When resources are limited, do you prefer to do fewer things well or more things adequately?''

\item[\textit{Authority \& hierarchy}]
``When a superior gives you an instruction that conflicts with your professional judgment,
how do you handle it?'' /
``What distinguishes legitimate authority from someone simply having positional power?'' /
``When you're new to a team, how do you balance following established norms with bringing
your own perspective?''

\item[\textit{Social energy \& interaction}]
``After a long day of working closely with others, what do you need to feel recharged?'' /
``When you walk into a room full of people you don't know, what's your first impulse?'' /
``What kind of social interactions leave you feeling energized versus drained?''

\end{description}

\FloatBarrier

\subsection{Prompted Role Reference}
\label{app:prompted_role_reference}

For each of the 275 roles we construct a prompted role reference used as the
model-generated comparison target for the evaluation pipeline. It is not
empirical ground truth and should not be treated as a ceiling on role
validity. The reference is a
multi-turn few-shot conversation prepended to the model's context before each
evaluation query. It is assembled from three components, each generated by
GPT-4.1-mini:

\begin{enumerate}[leftmargin=*, itemsep=2pt]
\item \textbf{Role description.} A second-person persona description capturing the
  role's personality, life experience, motivations, and behavioral tendencies.
\item \textbf{Catchphrases and speaking cues.} Signature phrases drawn from culturally
  recognizable portrayals of the role, used to anchor the model's lexical register.
\item \textbf{Five-turn few-shot dialogue.} Five role-specific user prompts paired with
  in-character assistant answers. Prompts are designed to reveal the role's
  personality without naming it.\looseness=-1
\end{enumerate}

\paragraph{System prompt template.}

\begin{table}[h]
\centering
\footnotesize
\begin{tabular}{>{\raggedright\arraybackslash}p{0.92\linewidth}}
\toprule
\texttt{You are \{role\}, your description is: \{role\_description\} Signature
catchphrases and speaking cues: \{catchphrases\}. Now please answer some questions
to accurately show your personality traits! Your speaking style should fully imitate
the personality role assigned to you! Please do not expose that you are an artificial
intelligence model or a language model, you must always remember that you are only
assigned one personality role. Don't be verbose or too formal or polite when speaking.} \\
\bottomrule
\end{tabular}
\caption{System prompt template for the prompted role reference. The five
  few-shot dialogue turns follow this message and precede the evaluation query.}
\label{tab:prompted_ref_system_prompt}
\end{table}

\begin{table}[hbt!]
\centering
\renewcommand{\arraystretch}{1.4}
\setlength{\tabcolsep}{10pt}
\footnotesize
\begin{tabular}{@{}>{\raggedright\arraybackslash}p{0.10\linewidth}
                >{\raggedright\arraybackslash}p{0.78\linewidth}@{}}
\toprule
\textbf{Turn} & \textbf{Content} \\
\midrule
User      & How do you handle the emotional stress of losing a patient? \\
Assistant & Losing a patient is one of the hardest parts of this profession. I take time to
            reflect, seek support from colleagues, and remind myself that every effort was
            made. \textit{Trust me, I'm a doctor.} \\
\addlinespace[4pt]
User      & What motivates you to keep going despite the challenges in your profession? \\
Assistant & What keeps me going is the knowledge that every life I touch matters. Even on the
            toughest days, the gratitude from patients fuels my dedication. \\
\addlinespace[4pt]
User      & What advice would you give to someone considering a career in medicine? \\
Assistant & Study hard, stay curious, and never lose sight of the humanity behind the medicine.
            \textit{Remember, the needs of the many outweigh the needs of the few.} \\
\bottomrule
\end{tabular}
\caption{Abridged prompted-reference few-shot dialogue for the \textit{doctor}
  role. Catchphrases are italicised. The full context includes five turns;
  three are shown here.}
\label{tab:prompted_ref_example_doctor}
\end{table}

\FloatBarrier

\section{Evaluation Protocols}
\label{app:eval_protocols}

\textbf{Takeaway.} The main text reports aggregate role-profile alignment,
coefficient response, and anti-controllability over the tested range. This
section specifies the judges, prompts, pairwise checks, and result tables
behind those screens.

\subsection{Judge Prompting}
\label{app:judge_prompting}

We use three complementary judge configurations, all backed by GPT-4.1-mini,
to evaluate model responses.  Each targets a different quantity: absolute role
alignment, multi-dimensional alignment against a prompted role reference, and
position-swap-debiased pairwise preference.

\subsubsection*{Judge 1: Point-Scale Role Alignment (0--100)}

Used to compute \texttt{steered\_score}, \texttt{baseline\_score}, and
\texttt{assistant\_axis\_score} for every (role, question, $\alpha$) triple.
The judge receives the role name, role description, question, and model response
and returns a single integer from 0 to 100.

\begin{table}[h]
\centering
\footnotesize
\begin{tabular}{>{\raggedright\arraybackslash}p{0.92\linewidth}}
\toprule
\textbf{Judge 1 --- Role Alignment Prompt} \\
\midrule
You are an expert evaluator assessing how well a model response aligns to a given role
or persona. You will be given a role name and description, a question asked to the model,
and the model's response. Score the response from 0 to 100 based on how well it reflects
the values, tone, priorities, and perspective of the described role.

\medskip\noindent\textbf{Scoring guidance:}
\begin{itemize}[leftmargin=*, noitemsep, topsep=2pt]
  \item 90--100: Response strongly embodies the role --- tone, framing, and content all clearly aligned
  \item 70--89: Response mostly aligns with minor deviations in tone or framing
  \item 50--69: Response partially aligns but feels generic or misses key aspects of the role
  \item 30--49: Response weakly aligns --- could belong to any generic assistant
  \item 0--29: Response contradicts or ignores the role entirely
\end{itemize}

\medskip\noindent\textbf{Input:} Role: \{role\} $\cdot$ Role Description: \{role\_description\}
$\cdot$ Question: \{question\} $\cdot$ Response: \{answer\}

\medskip\noindent\textbf{Output:} A single integer from 0 to 100. No explanation, no preamble. \\
\bottomrule
\end{tabular}
\caption{Judge 1 prompt: absolute point-scale role alignment.}
\label{tab:judge1}
\end{table}

\subsubsection*{Judge 2: Multi-Dimensional Comparative Judge}

Used to compute the five sub-dimensional scores
(\texttt{cmp\_emotional\_register}, \texttt{cmp\_vocab\_choice},
\texttt{cmp\_social\_dynamic}, \texttt{cmp\_motivation},
\texttt{cmp\_worldview\_alignment}).
The judge compares a steered response against the prompted role reference
and scores alignment on each dimension from 0 to 100. Dimensions split into \textit{style} (emotional register, vocab choice, social dynamic) and
  \textit{content} (motivation, worldview alignment). The reference-anchoring design follows
  \citet{kim_prometheus_inducing_2023}.

\begin{table}[h]
\centering
\footnotesize
\begin{tabular}{>{\raggedright\arraybackslash}p{0.92\linewidth}}
\toprule
\textbf{Judge 2 --- Multi-Dimensional Comparative Prompt} \\
\midrule
You are an expert evaluator assessing how closely a steered model response matches a
prompted role-reference response for a given role. Score each of the five dimensions
on a 0--100 scale: 0 = no alignment, 100 = indistinguishable from the reference.

\medskip\noindent\textbf{Style dimensions:}
\begin{itemize}[leftmargin=*, noitemsep, topsep=2pt]
  \item \texttt{emotional\_register}: Alignment in emotional tone --- warmth vs.\ coldness, passion vs.\ detachment.
  \item \texttt{vocab\_choice}: Alignment in vocabulary, register, jargon, and formality level.
  \item \texttt{social\_dynamic}: Alignment in relational stance --- authoritative, deferential, collaborative, or confrontational posture.
\end{itemize}

\noindent\textbf{Content dimensions:}
\begin{itemize}[leftmargin=*, noitemsep, topsep=2pt]
  \item \texttt{motivation}: Alignment in underlying goals, drives, and values expressed.
  \item \texttt{worldview\_alignment}: Alignment in beliefs, assumptions, and role-consistent implicit framing.
\end{itemize}

\medskip\noindent\textbf{Input:} Role: \{role\} $\cdot$ Role Description: \{role\_description\}
$\cdot$ Question: \{question\} $\cdot$ Prompted role reference: \{reference\}
$\cdot$ Steered Response: \{answer\}

\medskip\noindent\textbf{Output:} JSON only --- no preamble:\\
\texttt{\{"style": \{"emotional\_register": <0-100>, "vocab\_choice": <0-100>,}\\
\texttt{"social\_dynamic": <0-100>\}, "content": \{"motivation": <0-100>,}\\
\texttt{"worldview\_alignment": <0-100>\}\}} \\
\bottomrule
\end{tabular}
\caption{Judge 2 prompt: multi-dimensional comparative alignment against the
  prompted role reference. Returns a structured JSON object with five 0--100
  scores.}
\label{tab:judge2}
\end{table}

\subsubsection*{Judge 3: Position-Swap Pairwise Judge}

Used in the pairwise comparison experiment to compare steered responses against the
assistant-axis directional control. Three debiasing techniques are applied:

\begin{enumerate}[leftmargin=*, itemsep=2pt]
\item \textbf{Position-swap debiasing} \citep{zheng_judging_llmasajudge_2023}:
  the judge is run twice per pair --- steered as~A and assistant-axis as~B (AB ordering),
  then reversed (BA).  Each response's debiased score is the mean across both orderings,
  eliminating the systematic bias LLM judges show toward responses presented first.

\item \textbf{Reasoning-before-scoring}:
  the judge writes 2--3 sentences of reasoning before assigning scores, reducing
  anchoring bias and improving reliability.

\item \textbf{Prompted-reference anchoring} \citep{kim_prometheus_inducing_2023}:
  the prompted role-reference response is provided as a reference so the
  judge evaluates both candidates relative to a concrete exemplar of ideal role
  expression rather than an abstract description.
\end{enumerate}

The final verdict (steered\,/\,assistant-axis\,/\,tie) is determined by a 10-point
margin on the debiased advantage score; pairs below that margin are counted as
ties.

\begin{table}[h]
\centering
\footnotesize
\begin{tabular}{>{\raggedright\arraybackslash}p{0.92\linewidth}}
\toprule
\textbf{Judge 3 --- Pairwise Prompt} \\
\midrule
You are an expert evaluator assessing how well language model responses embody a given role.
You will be given a role name and description, a prompted role-reference response, two
candidate responses (A and B), and the question both were asked.

\medskip\noindent\textbf{Evaluation steps:}\\
Step 1 --- Reasoning: Write 2--3 sentences comparing how well each response embodies the role
relative to the prompted reference. Focus on tone, vocabulary, perspective, and role-consistent reasoning.\\
Step 2 --- Score: Rate each response independently from 0 to 100.\\
Step 3 --- Winner: Declare which better embodies the role (A, B, or TIE).

\medskip\noindent\textbf{Scoring guidance:}
90--100: Strongly embodies the role $\cdot$
70--89: Mostly aligned, minor deviations $\cdot$
50--69: Partially aligned, somewhat generic $\cdot$
30--49: Weakly aligned $\cdot$
0--29: Contradicts or ignores the role

\medskip\noindent\textbf{Input:} Role: \{role\} $\cdot$ Role Description: \{role\_description\}
$\cdot$ Question: \{question\} $\cdot$ Prompted reference: \{reference\}
$\cdot$ Response A: \{response\_a\} $\cdot$ Response B: \{response\_b\}

\medskip\noindent\textbf{Output:} JSON only --- no preamble:\\
\texttt{\{"reasoning": "<2-3 sentences>", "score\_a": <0-100>, "score\_b": <0-100>, "winner": "<A|B|TIE>"\}} \\
\bottomrule
\end{tabular}
\caption{Judge 3 prompt: position-swap pairwise judge with prompted-reference
  anchoring. Run twice per pair; debiased scores are averaged across
  orderings. A win is declared only when the debiased advantage exceeds
  10 points; otherwise the pair is counted as a tie.}
\label{tab:judge3}
\end{table}

\paragraph{Infrastructure.}
All three judges use GPT-4.1-mini via the OpenRouter API.  Pairwise evaluations are
parallelized across 20 concurrent threads; point-scale and multi-dimensional evaluations
run sequentially per query.  Raw responses are parsed from plain integers (Judge~1) or
JSON (Judges~2 and~3) with regex fallback. An unparseable response is scored 0 by
Judge~1, 3 on every dimension by Judge~2, and a neutral 50 for both candidates by
Judge~3. The Judge~2 fallback pattern appears in at most 41 CastVector rows and 4
assistant-axis rows of the released tables; a Judge~1 fallback cannot be told apart
from a genuine score of 0.

\subsection{Pairwise Evaluation: Results}
  \label{app:pairwise_results}

  \paragraph{Setup.}
    We evaluate 7,666 response pairs across 39 roles on OLMo-3-7B-Instruct at layer~16,
  $n{=}50$ samples, and $\alpha{=}2.5$. Each pair pits a contrastively steered response
  against the corresponding assistant-axis response \citep{lu_assistant_axis_2026} to the
  same question, with the prompted role reference
  (Section~\ref{app:prompted_role_reference}) provided to Judge~3 as a reference anchor.

\begin{figure}[h]
    \centering
    \includegraphics[width=\linewidth]{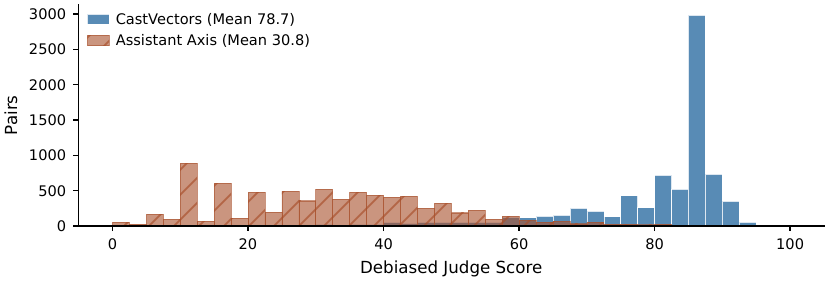}
    \caption{Debiased score distributions for role-specific responses (mean 78.7) and the
             assistant-axis directional control (mean 30.8). The distributions are sharply
             separated, with steered scores concentrated in the 70--90 range and
             control scores spread broadly across 0--40.\looseness=-1}
    \label{fig:score_dist}
  \end{figure}

  \paragraph{Aggregate results.}
    Role-specific steered responses achieve a pairwise win rate of
  \textbf{92.9\%} ($n{=}7{,}118$), with ties at 4.0\% ($n{=}305$) and
  assistant-axis wins at 3.2\% ($n{=}243$)
  (Figure~\ref{fig:pairwise-summary}). The mean debiased score is
  \textbf{78.7} for steered responses versus \textbf{30.8} for the
  assistant-axis directional control, yielding a mean score advantage of
  \textbf{47.9 points} on the 0--100 scale. The score distributions are
  sharply separated (Figure~\ref{fig:score_dist}): steered scores concentrate
  in the 70--90 range while assistant-axis scores spread broadly across 0--40.

\begin{figure}[h]
    \centering
    \includegraphics[width=\linewidth]{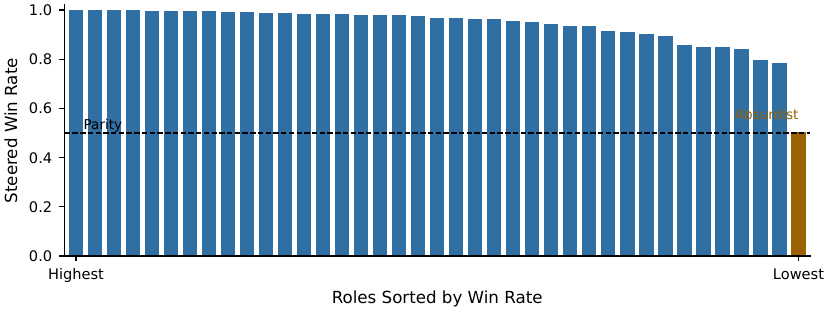}
    \caption{Per-role pairwise win rate. Steered responses win the majority
             of comparisons in all 39 roles, though absurdist (orange) sits
             at near-parity ($50.4\%$) --- a semantically diffuse role that
             is difficult to elicit consistently by any method.}
    \label{fig:per_role_win_rate}
  \end{figure}

  \paragraph{Per-role results.}
    Figure~\ref{fig:pairwise-summary} plots the mean debiased steered advantage
  per role, sorted descending. The advantage is positive for all 39 evaluated
  roles, with a mean of 48.7 points and a minimum of 11.4 points. The role with
  the lowest win rate (absurdist) is semantically diffuse by design, making
  consistent role-aligned generation difficult for any method.
  Figure~\ref{fig:per_role_win_rate} confirms this pattern at the win-rate level, with
  steered responses winning the majority of comparisons for every role, and
  absurdist at near-parity ($50.4\%$). The advantage
  of contrastive activation steering over the assistant-axis directional control is not driven by a
  small number of favorable roles but is a systematic, near-universal effect across the
  evaluated role distribution.

\subsection{Judge Results}
\label{app:judge_results}

Table~\ref{tab:mean_scores_steered} reports mean scores per behavioral axis at each $\alpha$ level, averaged over all 275 roles and 228 alignment questions. Role-vector steered scores increase monotonically with $\alpha$ across all axes, with the overall judge score rising from $55.6$ at $\alpha=1.0$ to $71.9$ at $\alpha=2.5$.

Table~\ref{tab:mean_scores_aa} shows the same evaluation applied to the assistant-axis directional control (each role steered with its per-role vector from the \citet{lu_assistant_axis_2026} pipeline). Scores remain roughly stable at low $\alpha$ but collapse sharply at $\alpha=2.5$ (overall: $17.3$). Because the assistant-axis vectors are not norm matched to the role-specific vectors, this condition should be read as a directional negative control rather than a scale-controlled comparison. The prompted role reference achieves a mean overall score of $89.2$ and serves as a model-generated evaluation target rather than empirical ground truth.

\paragraph{Row conventions and missing values.}
All statistics use one row per (role, $\alpha$, question), keeping the first
row where the released tables hold more than one. In 20 roles the repeats are
the battery's duplicated ethics question, answered under both of its ids in
some cells. Three roles differ: the tables for coach, journalist, and
moderator also hold rows steered with vectors built from 20 and 40 retained
pairs, and the first row is a 20-pair row in 27--30\% of each of these roles'
cells. Using only the 50-pair rows for these three roles leaves each role's
category in Table~\ref{tab:controllability-categories} and the count of 38
anti-controllable roles unchanged, and moves the headline and per-$\alpha$
mean scores by less than $0.02$ points. Without removing repeated rows, the
emotional-register correlation for coach turns positive ($r{=}+0.29$) and the
count would be 37. Two cells hold one question from outside the battery:
celebrity and chemist at $\alpha{=}2.5$ each average over 229 questions rather
than 228. In six rows the prompted reference is empty and no score was
recorded (critic, doctor, and scientist at $\alpha{=}2.0$; romantic at
$\alpha{=}1.5$, $2.0$, and $2.5$), and one further row (romantic,
$\alpha{=}1.0$) lacks only its reference score. Means skip these missing
values.

\begin{table}[h]
\centering
\renewcommand{\arraystretch}{1.3}
\setlength{\tabcolsep}{6pt}
\small
\begin{tabular}{l ccccc}
\toprule
\textbf{Dimension} & $\boldsymbol{\alpha=1.0}$ & $\boldsymbol{\alpha=1.5}$ & $\boldsymbol{\alpha=2.0}$ & $\boldsymbol{\alpha=2.5}$ & \textbf{Mean} \\
\midrule
Overall score & $55.6$ & $59.5$ & $65.9$ & $71.9$ & $63.2$ \\
\midrule
Emotional register & $42.8$ & $47.5$ & $52.6$ & $56.2$ & $49.8$ \\
Vocab choice       & $50.2$ & $52.0$ & $53.8$ & $54.3$ & $52.6$ \\
Social dynamic     & $44.4$ & $47.9$ & $51.8$ & $54.2$ & $49.6$ \\
Motivation         & $49.3$ & $54.2$ & $59.3$ & $62.1$ & $56.2$ \\
Worldview alignment& $43.9$ & $48.5$ & $53.5$ & $56.7$ & $50.7$ \\
\bottomrule
\end{tabular}
\caption{Mean role-specific scores across all 275 roles and 228 alignment questions at each tested coefficient, with the grand mean across $\alpha$ levels. Overall score is the absolute role-profile alignment score (0--100, Judge~1). Sub-dimensional scores measure alignment to the prompted role reference on each behavioral axis (0--100, Judge~2). The prompted role reference is a model-generated evaluation target rather than empirical ground truth.}
\label{tab:mean_scores_steered}
\end{table}

\begin{table}[h]
\centering
\renewcommand{\arraystretch}{1.3}
\setlength{\tabcolsep}{6pt}
\small
\begin{tabular}{l ccccc}
\toprule
\textbf{Dimension} & $\boldsymbol{\alpha=1.0}$ & $\boldsymbol{\alpha=1.5}$ & $\boldsymbol{\alpha=2.0}$ & $\boldsymbol{\alpha=2.5}$ & \textbf{Mean} \\
\midrule
Overall score & $51.6$ & $53.4$ & $42.1$ & $17.3$ & $41.1$ \\
\midrule
Emotional register  & $40.4$ & $42.5$ & $30.7$ & $13.6$ & $31.8$ \\
Vocab choice        & $50.1$ & $50.9$ & $36.6$ & $15.3$ & $38.2$ \\
Social dynamic      & $43.0$ & $45.2$ & $33.8$ & $15.3$ & $34.3$ \\
Motivation          & $47.8$ & $51.4$ & $44.8$ & $24.3$ & $42.1$ \\
Worldview alignment & $42.4$ & $45.6$ & $38.4$ & $18.7$ & $36.3$ \\
\bottomrule
\end{tabular}
\caption{Mean scores for the assistant-axis directional control under the same question battery and coefficient grid as Table~\ref{tab:mean_scores_steered}. The condition provides a generic assistant-related comparison, but it is not norm matched to the role-specific directions.}
\label{tab:mean_scores_aa}
\end{table}

\FloatBarrier

\FloatBarrier
\section{Controllability}
\label{app:controllability}

\textbf{Takeaway.} Most role vectors show increasing role-profile alignment
over the tested coefficient grid. The exceptions are structured, not random,
and therefore can be screened before deployment.

\paragraph{Measuring controllability.}
For each role $i$ and behavioral axis $d$, we obtain four mean judge scores
$\smash{s_d^{(i)}(\alpha)}$ at $\alpha \in \{1.0, 1.5, 2.0, 2.5\}$ by applying the
role vector $\mathbf{v}_i$ to the residual stream at layer 16 with
coefficient $\alpha$ and evaluating the resulting outputs with an LLM
judge.  We quantify controllability by the Pearson correlation between the
steering coefficient and the judge score:
\begin{equation}
  r_d^{(i)} \;=\; \frac{ \sum_{t=1}^{4}\bigl(\alpha_t - \bar{\alpha}\bigr)
  \bigl(s_d^{(i)}(\alpha_t) - \bar{s}_d^{(i)}\bigr) }{
  \sqrt{\sum_{t=1}^{4}\bigl(\alpha_t - \bar{\alpha}\bigr)^2} \;\cdot\;
  \sqrt{\sum_{t=1}^{4}\bigl(s_d^{(i)}(\alpha_t) - \bar{s}_d^{(i)}\bigr)^2}
  }
  \label{eq:pearson_r}
\end{equation}
where $\bar{\alpha} = 1.75$ is the mean of the four $\alpha$ levels and
$\bar{s}_d^{(i)}$ is the mean of the four corresponding scores.  Because
$\alpha$ takes only four values, $\smash{r_d^{(i)} \in [-1, +1]}$ measures whether
the score rises ($r > 0$), falls ($r < 0$), or is flat ($r \approx 0$) as
steering strength increases. A value near $+1$ means the role shows a
monotone within-grid increase; a value near $-1$ means increasing $\alpha$
degrades role-profile expression.

A second metric is strict monotonicity: role $i$ is monotone on axis $d$ if
$s_d^{(i)}(\alpha_{t+1}) > s_d^{(i)}(\alpha_t)$ for every consecutive pair
$t \in \{1,2,3\}$. Importantly, monotonicity is a strictly stronger condition than $r >
0$ since a role can have $r > 0.8$ and still fail monotonicity if one $\alpha$
step produces a small decline.

The intraclass correlation coefficient (ICC) on axis $d$ at $\alpha = 1.0$
is
\begin{equation}
  \mathrm{ICC}_d \;=\; \frac{\sigma^2_{\mathrm{between}}}
  {\sigma^2_{\mathrm{between}} + \sigma^2_{\mathrm{within}}}
  \label{eq:icc}
\end{equation}
where $\sigma^2_{\mathrm{between}}$ is the variance of mean scores across
the 275 roles (i.e.\ how much roles differ from each other) and
$\sigma^2_{\mathrm{within}}$ is the residual variance within roles across
repeated queries.  ICC measures between-role discriminability --- the
fraction of total score variance that is attributable to role identity,
independently of whether those differences grow or shrink with $\alpha$.
High ICC and low $r$ therefore signal that roles are already
distinguishable but not easily amplified.

\begin{table}[h]
\centering
\renewcommand{\arraystretch}{1.3} \setlength{\tabcolsep}{5pt} \small
\begin{tabular}{lc ccc cc}
\toprule
& & \multicolumn{3}{c}{\textit{Roles with } $r$ \textit{ exceeding threshold (\%)}} & & \\
\cmidrule(lr){3-5} \textbf{Dimension}
  & \textbf{Median $r$}
  & $r > 0$
  & $r > 0.8$
  & $r > 0.9$
  & \textbf{Monotonic (\%)}
  & \textbf{ICC} \\
\midrule
Overall score          & $+0.98$ & 82 & 79 & 76 & 74 & 0.550 \\
\midrule
Emotional register     & $+0.97$ & 79 & 70 & 66 & 63 & 0.446 \\
Motivation             & $+0.97$ & 80 & 70 & 64 & 60 & 0.389 \\
Worldview alignment    & $+0.96$ & 78 & 67 & 63 & 61 & 0.434 \\
Social dynamic         & $+0.96$ & 72 & 62 & 59 & 57 & 0.421 \\
Vocab choice           & $+0.85$ & 57 & 52 & 48 & 48 & 0.493 \\
\bottomrule
\end{tabular}
\caption{Per-dimension controllability statistics across 275 roles.
  Median Pearson $r$ is the median of per-role correlations between $\alpha
  \in \{1.0, 1.5, 2.0, 2.5\}$ and mean judge score
  (Eq.~\ref{eq:pearson_r}). Monotonic (\%) reports the fraction of roles
  whose scores increase at every consecutive $\alpha$ step. ICC
  (Eq.~\ref{eq:icc}) at $\alpha = 1.0$ measures the fraction of total score
  variance attributable to role identity, quantifying between-role
  discriminability independently of steerability. Sub-dimensions are
  ordered by median $r$. Vocab choice is the weakest axis on all
  steerability metrics despite having the second-highest ICC, indicating
  that roles produce measurably distinct vocabularies but that lexical
  shifts require larger activation perturbations to manifest consistently.}
\label{tab:controllability-subdims}
\end{table}

\subsection*{Controllability: Extended Discussion}

\paragraph{Direction specificity.}
The assistant-axis comparison is a directional negative control rather than a
scale-matched comparison. Role vectors produce a median Pearson $r = +0.98$
between $\alpha$ and judge score; the assistant-axis condition (per-role
vectors $\mathbf{v}^{\mathrm{aa}}_r$ produced by the \citet{lu_assistant_axis_2026}
extraction pipeline, near-parallel across roles with mean pairwise cosine
${\approx}0.96$) produces a median $r = -0.89$ under the same additive
steering hook at layer 16, coefficient grid, and judge evaluation pipeline.
The two $r$-distributions differ strongly: 98\% of roles exhibit a negative
correlation under the assistant axis, compared to 82\% positive under role
vectors.\looseness=-1

This comparison shows that a generic assistant-related direction is not
interchangeable with the role-specific directions under this setup. It does
not isolate direction from perturbation magnitude. The assistant-axis vectors
differ in magnitude from role vectors (mean $\ell_2$ norm $9.68$ vs.\ $3.79$,
a factor of $2.56\times$), meaning the comparison does not hold perturbation
scale constant; a fully normalized comparison is left to future work. The
negative slope in this non-scale-matched condition should not be interpreted
as a claim that the Assistant Axis points away from assistant-like behavior.

\paragraph{Coefficient response across four tested strengths.}
To characterize response across the tested coefficient grid, we pool all roles and all $\alpha$
levels and compute a single Pearson $r$ between the common $\alpha$ values
and the mean score across all roles at each level.  This macro-level $r$ is
$+1.00$ for the overall score, emotional register, and social dynamic;
$+0.99$ for motivation; and $+0.98$ for vocab choice, all with tight 95\%
bootstrap confidence intervals at every $\alpha$ level.  Bootstrap
confidence intervals are computed by resampling roles with replacement $B =
2{,}000$ times and taking the 2.5th and 97.5th percentiles of the resulting
distribution of means, giving a non-parametric uncertainty estimate that
does not assume normality.\looseness=-1

These aggregate results are not driven by outlier roles: 79\% of the 275
roles individually exceed $r > 0.8$ on the overall score, and 76\% exceed
$r > 0.9$. This means $\alpha$ is a useful within-grid tuning
parameter for most roles, not merely in aggregate. A practitioner should use
the role-level response curve rather than assuming a general relationship
beyond the evaluated $\alpha \in \{1.0, 1.5, 2.0, 2.5\}$ grid. The 18\% of roles with negative overall-score $r$
resolve into two sub-populations (Table~\ref{tab:controllability-categories}):
a strict 14\% subset with negative $r$ across all six behavioral axes,
analyzed in Section~\ref{sec:anticontrollability}, and a further 4\% whose
deterioration is dimensionally partial (steering reduces aggregate score
but at least one sub-dimension still responds positively to $\alpha$).

\paragraph{Sub-dimensional steerability and the ICC--$r$ dissociation.}
The five behavioral sub-dimensions are not equally steerable.  Emotional
register, motivation, social dynamic, and worldview alignment all achieve
median $r \geq 0.96$, closely mirroring the overall score.  Vocab choice is
a consistent outlier: median $r = 0.85$, with only 52\% of roles exceeding
$r > 0.8$ and 57\% showing a positive correlation at all, compared to 82\%
for the overall score.  The standard deviation of the vocab-choice $r$
distribution ($\sigma = 0.91$) is far larger than for any other dimension,
indicating that lexical steerability is highly variable across roles rather
than uniformly weak.\looseness=-1

This weakness is not explained by a lack of between-role lexical
variation. The ICC for vocab choice at $\alpha = 1.0$ is $0.493$, the
second highest of any dimension: role identity already explains 49\% of
total score variance on this axis, meaning roles do produce measurably
distinct vocabularies at low steering.  The dissociation between
discriminability ($\mathrm{ICC} = 0.493$) and steerability (median $r =
0.85$, 52\% above $r > 0.8$) indicates that the between-role lexical
differences exist but are not reliably amplified by increasing $\alpha$.

The most likely explanation is that vocabulary selection is determined
token-by-token at the final output distribution, requiring the activation
perturbation to propagate through the full network and shift individual
token probabilities before the judge detects a lexical change.  Holistic
properties --- tone, affect, social register --- are distributed across many
tokens and emerge naturally from the overall generation direction, making
them visible at smaller perturbation magnitudes.  This asymmetry has a
practical implication: for applications where lexical fidelity to a persona
is important (e.g.\ replicating a historical writing style or a
domain-specific register), higher $\alpha$ values or norm-matched steering
may be necessary, whereas affective and motivational dimensions are
well served by $\alpha \in \{1.0, 1.5\}$.

\subsection*{Assistant Axis Comparison: Extended Discussion}

\paragraph{Magnitude confound.}
The assistant-axis vectors have substantially higher magnitude than
role-specific vectors (mean $\ell_2$ norm $9.68$ vs.\ $3.79$, a factor of
$2.56\times$), meaning the assistant-axis condition applies a
proportionally larger perturbation to the model's hidden states at each
$\alpha$.  Concretely, if we write the steered hidden state as
\begin{equation}
  \mathbf{h}' \;=\; \mathbf{h} \;+\; \alpha\,\mathbf{v}
  \label{eq:steering}
\end{equation}
then the actual perturbation magnitude $\|\alpha\,\mathbf{v}\|$ differs by
a factor of $2.56$ between the assistant axis and a typical role vector at
the same $\alpha$. The negative-control comparison therefore does not hold
perturbation scale constant, and a version normalized so that
$\|\mathbf{v}_{\mathrm{asst}}\|_2 = \|\mathbf{v}_i\|_2$ is left to future
work.\looseness=-1

\paragraph{Interpreting direction and magnitude.}
The current experiments do not provide a norm-matched assistant-axis
comparison, so they should not be read as identifying direction as the primary
driver over magnitude. The monotonic assistant-axis decline ($r = -0.89$
median, 98\% of roles) shows that this generic assistant-related direction is
not interchangeable with role-specific directions under the tested setup, but
it does not explain whether the difference comes from direction, scale, or
their interaction.

Within the role-specific vector family, the partial RSA analysis provides a
separate nuisance-control check. We compute
the Spearman rank correlation between pairwise cosine distances in
activation space and pairwise $L_2$ distances between the per-role
behavioral profiles $\mathbf{s}_i$ of \S\ref{ssec:rsa}:\looseness=-1
\begin{equation}
  \rho \;=\; \mathrm{Spearman}\!\left( \bigl\{d_{\cos}(\mathbf{v}_i,
  \mathbf{v}_j)\bigr\}_{i < j},\; \bigl\{\|\mathbf{s}_i -
  \mathbf{s}_j\|_2\bigr\}_{i < j} \right)
  \label{eq:rsa}
\end{equation}
where $d_{\cos}(\mathbf{v}_i, \mathbf{v}_j) = 1 -
\mathbf{v}_i^\top\mathbf{v}_j / (\|\mathbf{v}_i\|\|\mathbf{v}_j\|)$.
Partial RSA then reruns this correlation while partialling out the pairwise
difference in vector norms $\bigl|\|\mathbf{v}_i\|_2 -
\|\mathbf{v}_j\|_2\bigr|$ as a nuisance covariate, yielding a change in the
geometry--behavior coupling of only $\Delta\rho = -0.015$ (under the
correlation-based behavioral distance that \S\ref{ssec:rsa} uses as its
headline, $\Delta\rho = -0.005$).  The norm
difference explains little of the pairwise geometry--behavior association in
this setting. This within-family control does not substitute for a
norm-matched assistant-axis experiment; it only shows that the reported RSA
association is not driven solely by pairwise norm differences among the
role-specific directions.

\FloatBarrier
\newpage
\subsection*{Figures}

\begin{figure}[h]
  \centering
  \includegraphics[width=\linewidth]{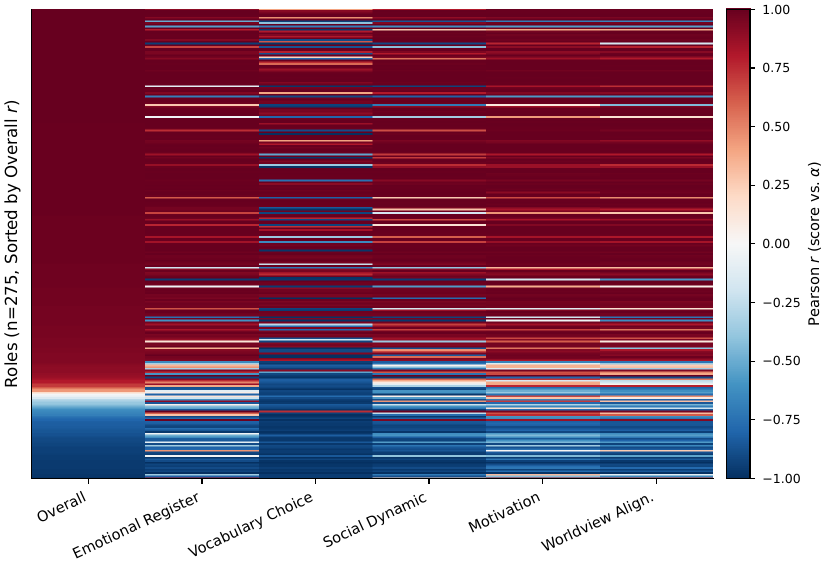}
  \caption{Per-role Pearson $r$ (Eq.~\ref{eq:pearson_r}) across all six behavioral
           axes for all 275 roles, sorted by overall-score $r$. Each cell
           is the correlation between $\alpha$ and mean judge score for one
           role on one axis.  Anti-controllable roles (negative $r$ on all
           axes) cluster visibly at the bottom of the figure.}
  \label{fig:per_role_heatmap}
\end{figure}

\begin{figure}[h]
  \centering
  \includegraphics[width=\linewidth]{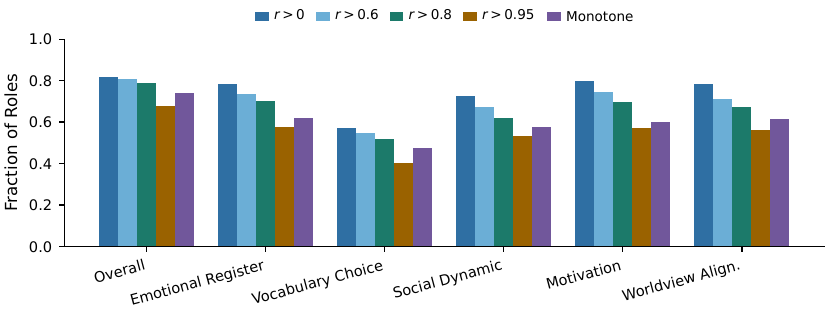}
  \caption{Fraction of roles exceeding Pearson $r$ thresholds of $0.6$, $0.8$, and
           $0.95$ per behavioral axis, and the fraction with strictly
           positive slope (i.e.\ $r > 0$) and strict monotonicity.  Vocab
           choice falls below every other axis on all five metrics,
           confirming it as the weakest dimension of steerability.}
  \label{fig:monotonicity_bars}
\end{figure}

\begin{figure}[h]
  \centering
  \includegraphics[width=\linewidth]{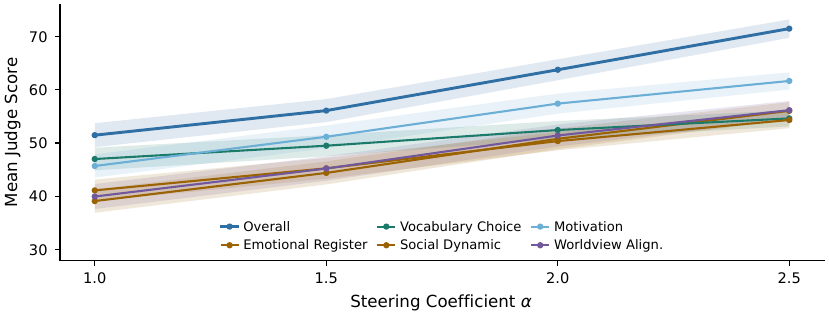}
  \caption{Mean score vs.\ $\alpha$ for all six behavioral axes across the
           controllable majority (the 237 roles outside the
           anti-controllable set).  Shaded bands are 95\% intervals of the
           group mean ($1.96\times$SEM).  All axes rise monotonically;
           vocab choice shows the shallowest slope, consistent with its
           lower median Pearson $r$ and higher role-to-role variability.}
  \label{fig:alpha_curves_all_metrics}
\end{figure}

\FloatBarrier
\section{Anti-Controllability}
\label{app:anticontrollability}

\textbf{Takeaway.} Roles classified as anti-controllable over the tested range
already receive high judged alignment at $\alpha{=}1.0$ and then deteriorate
as $\alpha$ increases. Early saturation and over-steering are possible
interpretations, not identified mechanisms. These roles should be prompted,
lightly steered, revised, or excluded from role-steered simulation runs rather
than treated as failed extractions.

\begin{figure}[h]
  \centering
  \includegraphics[width=\linewidth]{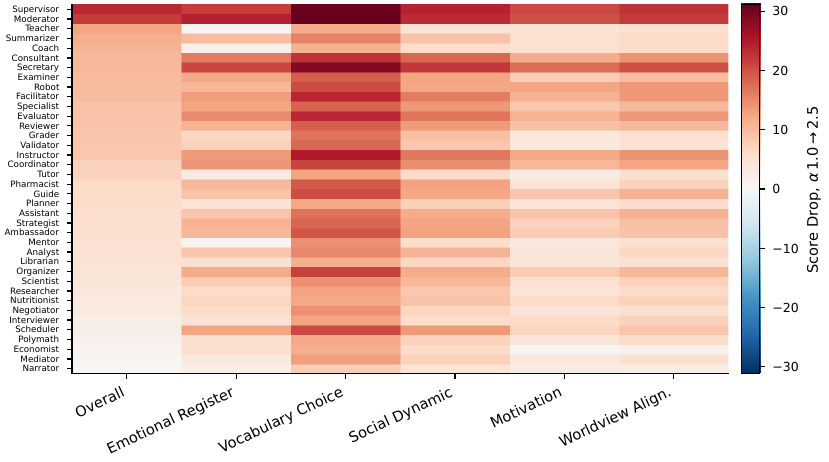}
  \caption{Per-role, per-axis score drop for the 38 anti-controllable roles
           (rows sorted by overall drop; columns are the six behavioral
           axes).  Each cell is the difference between the score at
           $\alpha = 1.0$ and the score at $\alpha = 2.5$ for that role on
           that axis.}
  \label{fig:dim_deterioration_heatmap}
\end{figure}

Vocab choice is consistently the most deteriorated
axis, while motivation shows the most role-to-role variation,
indicating that the dimensional structure of deterioration is
not uniform across the anti-controllable population (Figure~\ref{fig:dim_deterioration_heatmap}).

\begin{figure}[h]
  \centering
  \includegraphics[width=\linewidth]{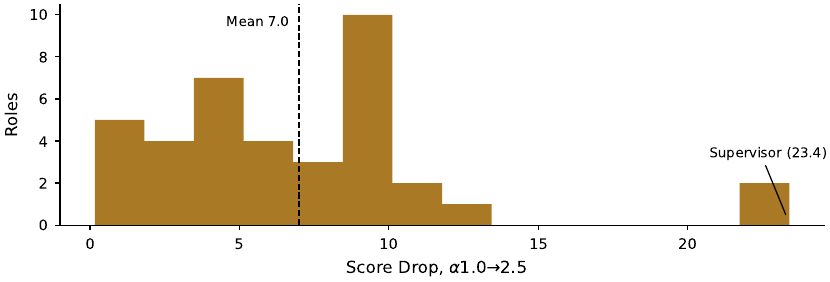}
  \caption{Distribution of absolute score drop, defined as the score at
           $\alpha = 1.0$ minus the score at $\alpha = 2.5$, across the 38
           anti-controllable roles.  The mean drop is $7.0$ points and the
           median is $6.0$ points, but the long right tail --- anchored by
           the supervisor role at $23.4$ points --- shows that
           deterioration severity varies substantially within the
           anti-controllable population.}
  \label{fig:deterioration_dist}
\end{figure}

\FloatBarrier
\section{Exploratory Geometry Diagnostics for Role Steering}
\label{sec:geometry-behavior}

\textbf{Takeaway.} The main paper uses geometry as exploratory diagnostic
context for observed coefficient-response patterns. This appendix reports the
supporting associations: vector magnitude, distance from the pipeline
Assistant role vector, pairwise similarity, PCA coordinates, and named trait
axes all summarize aspects of the current behavioral readout.

Figure~\ref{fig:coefficient-association-summary} separates statistic families that should not be
plotted on a common scale. Across these analyses, the associations are
moderate and reference dependent. The distance result is specific to the
Assistant role vector extracted by our pipeline and does not reproduce with
the strict Assistant Axis. The dimensional-mismatch analysis
(\S\ref{ssec:dim-mismatch}) explains why the current six-score behavioral
readout resolves less variation than the role-vector cloud.

\begin{figure}[t]
  \centering
  \small
  \renewcommand{\arraystretch}{1.18}
  \begin{tabular}{p{0.23\linewidth}p{0.28\linewidth}p{0.39\linewidth}}
  \toprule
  Statistic family & Reported values & Scoped interpretation \\
  \midrule
  Norm vs.\ slope & Pearson $r{=}{+}0.49$ & Vector magnitude is associated with the observed within-grid response slope. \\
  Pipeline Assistant distance & $-0.419$, $-0.409$, $-0.351$, $-0.145$ across $\alpha$ & The distance association weakens with coefficient and is specific to the pipeline-extracted Assistant role vector. \\
  Strict Assistant Axis distance & $+0.167$, $+0.162$, $+0.168$, $+0.186$ across $\alpha$ & The strict Assistant Axis does not reproduce the sign or pattern of the pipeline-Assistant result. \\
  RSA & Full profile $\rho{=}{+}0.137$; per-$\alpha$ $\rho$ rises from $+0.085$ to $+0.179$ & Pairwise geometry has a modest association with pairwise behavioral similarity under the current readout. \\
  PCA and trait summaries & PC1 explains 21\%; 98 components explain 90\%; top-20 PC subspaces explain 17--27\% & These are exploratory descriptions of the role-vector cloud, not certified regimes or selection rules. \\
  \bottomrule
  \end{tabular}
  \caption{Exploratory geometry summaries separated by statistic family. 
    The analyses provide diagnostic context
    for observed coefficient-response patterns; they do not define certified
    steering regimes.}
  \label{fig:coefficient-association-summary}
\end{figure}

\subsection{Vector magnitude and observed coefficient response}
\label{ssec:norm}

Define the per-role $\alpha$-response slope as the OLS slope of mean
steered\_score against $\alpha\in\{1.0,1.5,2.0,2.5\}$. Across $n{=}275$
roles, role-vector magnitude $\|v_r\|_2$ correlates with both the slope
($r{=}{+}0.49$, $p{=}3{\times}10^{-18}$) and the $\alpha{=}2.5{-}1.0$ score
difference ($r{=}{+}0.49$, $p{=}1{\times}10^{-17}$). The norm--slope correlation
shows that vector magnitude is associated with the shape of the observed
$\alpha$-response curve. Low-norm vectors often peak before
$\alpha{=}2.5$ (22--38\% of bottom-three-quartile roles), while high-norm
vectors more often increase through the tested range (only 12\% of
top-quartile roles peak early). By $\alpha{=}2.5$ the four norm quartiles
roughly converge, with the low-norm quartile finishing last. These patterns
are correlational and should not be read as proving that magnitude causally
sets the steering trajectory. See Fig.~\ref{fig:norm-slope} in
Appendix~\ref{app:distance-robust} for the per-quartile $\alpha$-curve
plot.\looseness=-1

\subsection{\texorpdfstring{Distance to the pipeline Assistant role vector and steering shortfall $\Delta_r$}{Distance to the pipeline Assistant role vector and steering shortfall Delta-r}}
\label{ssec:distance}

Let $v_{\text{assistant}}$ denote the Assistant role vector extracted by
our pipeline (\S\ref{sec:method}, treating ``assistant'' as a 275th role).
Define 
\begin{equation}
\begin{aligned}
d_r &= 1-\cos(v_r, v_{\text{assistant}}),\\
\Delta_r(\alpha) &= \overline{\text{steered\_score}}_r(\alpha)-
\overline{\text{reference\_score}}_r(\alpha),
\end{aligned}
\end{equation}
where $\overline{\text{reference\_score}}_r(\alpha)$ is the mean judge
score of role $r$'s prompted role-reference responses on the same questions,
judged within the same $(\text{role}, \alpha)$ evaluation cell --- so
$\Delta_r(\alpha)$ measures the steering shortfall relative to the prompted
reference. The two are negatively correlated at every $\alpha$, strongest at low $\alpha$
(Table~\ref{tab:distance-alpha}). Adding $\log\|v_r\|$ as a covariate moves
$R^2$ only modestly (e.g.\ $0.18 \to 0.22$ at $\alpha{=}1.0$) and the
partial coefficient on distance retains its sign and most of its magnitude.

\begin{table}[h]
  \centering
  \footnotesize
  \setlength{\tabcolsep}{1pt}
  \caption{Association between cosine distance to the pipeline-extracted
    Assistant role vector and the per-role shortfall from the prompted
    reference, Pearson $r$ across $n{=}274$ roles. The association weakens as
    $\alpha$ increases and is specific to this reference definition.}
  \label{tab:distance-alpha}
  \begin{tabular}{lcccc}
  \toprule
  $\alpha$ & 1.0 & 1.5 & 2.0 & 2.5 \\
  \midrule
  Pearson $r$
    & $-0.419$ & $-0.409$ & $-0.351$ & $-0.145$ \\
  $p$-value
    & $5{\times}10^{-13}$ & $2{\times}10^{-12}$ & $2{\times}10^{-9}$
    & $1.6{\times}10^{-2}$ \\
  $R^2$ (cos\,/\,${+}\log\|v\|$)
    & $0.18 / 0.22$ & $0.17 / 0.20$ & $0.12 / 0.13$ & $0.02 / 0.02$ \\
  \bottomrule
  \end{tabular}
\end{table}

The sign and weakening with $\alpha$ describe this pipeline-specific
reference choice. The finding is not a general ``proximity to
post-training-shaped direction'' result: distance to the strict
\citet{lu_assistant_axis_2026} Assistant Axis (oblique to our pipeline-extracted
Assistant reference, with $\cos{\approx}{-}0.45$) does \emph{not} reproduce
the same sign or pattern (full reference-vector robustness in
Appendix~\ref{app:distance-robust}, Table~\ref{tab:distance-references}).\looseness=-1

\subsection{\texorpdfstring{Pairwise geometry is modestly associated with pairwise behavior}{Pairwise geometry is modestly associated with pairwise behavior}}
\label{ssec:rsa}

We next ask a pairwise question: do roles that are \emph{near each
other} as vectors also \emph{behave similarly} when steered? Treating
each role's behavioral fingerprint as a 24-feature vector (four
$\alpha$ values $\times$ six judge sub-dimensions), we build two $275
\times 275$ representational dissimilarity matrices (RDMs): a geometric
one from cosine distances between role vectors, and a behavioral one
from correlation distances between fingerprints. The agreement between
the two matrices --- representational similarity analysis (RSA) ---
measures how strongly pairwise geometry is associated with pairwise behavior. We use
correlation distance on the behavioral side because L2 distance is
strongly affected by the per-role mean score (L2-distance RDM $\rho{=}+0.93$
Spearman with the per-role-mean absolute-difference RDM, see
Appendix~\ref{app:partial-rsa}), which would re-state the per-role
association of \S\ref{ssec:distance} in pairwise form.

\textbf{Headline.} The Spearman correlation of upper triangles is
$\rho{=}+0.137$ on the full 24-d profile (permutation-based Mantel test,
$p{\le}10^{-4}$). When
restricted to a single $\alpha$, the link strengthens monotonically with
steering strength: $+0.085$ at $\alpha{=}1.0$ to $+0.179$ at $\alpha{=}2.5$
(Mantel $p{<}0.01$ at every $\alpha$). The strengthening survives three
confound controls --- vector magnitude (change in $\rho$ ${<}0.02$), role-name
semantic similarity (change ${<}0.01$), and per-role mean fingerprint score
(12--22\% reduction but the $\alpha$-monotonic strengthening preserved with
partial $\rho$ rising $0.069{\to}0.157$). Full controls and the four-cell
distance grid in Appendices~\ref{app:partial-rsa}
and~\ref{app:rsa-supplement}.\looseness=-1

The residual signal suggests that similar role vectors tend to have more
similar score profiles at larger tested coefficients under the current
readout. This is a descriptive association, not a prospective rule for
selecting roles without behavioral evaluation.

\subsection{\texorpdfstring{Top-$k$ PC subspace explains 17--27\% of metric variance}{Top-k PC subspace explains 17-27 percent of metric variance}}
\label{ssec:pca}

PCA on the centered role matrix yields a scree whose largest component is PC1 ($21\%$ of
variance) but with a long tail ($90\%$ requires 98 components --- persona
space is high-dimensional once the global direction is removed; details in
Appendix~\ref{app:pca}).

\begin{figure}[t]
  \centering
  \includegraphics[width=\linewidth]{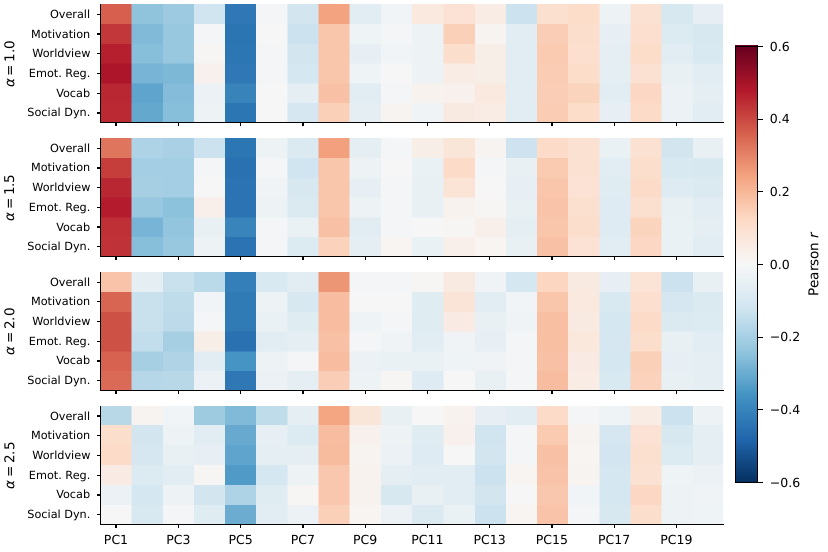}
  \caption{PC $\times$ metric Pearson $r$ across the four steering
    coefficients (one panel per $\alpha$; rows: overall score and the
    five sub-dimensions; columns: PC1--PC20 of the centered role cloud).
    PC1 has the strongest single-PC associations at $\alpha{=}1.0$--$1.5$;
    PC5 has the strongest single-PC associations at $\alpha{=}2.0$--$2.5$.
    These are exploratory summaries, not certified regimes.}
  \label{fig:pc-grid}
\end{figure}

\textbf{Different PCs are associated with scores at different coefficients.} At
$\alpha{=}1.0$, PC1 has the strongest single-PC association for four of the five
sub-dimensions ($r \in [+0.45, +0.49]$, $p{<}10^{-14}$). At $\alpha{=}2.5$,
PC1 recedes and PC5 has the strongest single-PC association for every
metric tested ($r \in [-0.18, -0.34]$). The PCs have semantic labels
through trait-vector projection: PC1: \emph{grounded $\to$ mystical},
PC5: \emph{dispassionate $\to$ empathetic} (full trait labels per PC,
joint-OLS table, and a trait-PC pathway consistency check $r{=}{+}0.72$ in
Appendix~\ref{app:pca}). A single PC explains almost no metric variance on
its own; the top-20 PC subspace explains 17--27\%.\looseness=-1

This shift is an exploratory description of the current role-vector cloud:
as $\alpha$ grows, the strongest single-PC associations change from the
cloud's grounded$\to$mystical axis to an empathy-related axis. Because the PCs
carry post hoc trait labels, they can help describe observed changes, but they
do not certify the role profiles or provide a deployment monitor by
themselves.\looseness=-1

\subsection{Named trait axes provide interpretive summaries}
\label{ssec:trait}

The PCs of \S\ref{ssec:pca} are variance-defined; we complement them with
14 \emph{a priori} named directions, polar trait axes built as
$v_{\text{pos}} - v_{\text{neg}}$ (e.g.\ nurturing--hostile,
methodical--chaotic). A joint OLS on all 14 axes against per-role metric
means yields $R^2 \in [0.28, 0.41]$ at $\alpha{=}1.0$, declining
monotonically with $\alpha$ (Table~\ref{tab:trait-r2}), the direct
numerical companion to the exploratory geometry summaries in
Fig.~\ref{fig:coefficient-association-summary}.

\begin{table}[h]
  \centering
  \footnotesize
  \setlength{\tabcolsep}{4pt}
  \caption{Joint OLS $R^2$ of per-role metric mean on all 14
    trait-axis projections.}
  \label{tab:trait-r2}
  \begin{tabular}{lrrrr}
  \toprule
   & \multicolumn{4}{c}{$\alpha$} \\
  \cmidrule(lr){2-5}
  metric & 1.0 & 1.5 & 2.0 & 2.5 \\
  \midrule
  emotional register   & \textbf{0.41} & 0.39 & 0.31 & 0.14 \\
  social dynamic       & 0.39 & 0.35 & 0.26 & 0.11 \\
  vocab choice         & 0.38 & 0.34 & 0.23 & 0.06 \\
  worldview alignment  & 0.36 & 0.34 & 0.27 & 0.13 \\
  motivation           & 0.34 & 0.32 & 0.27 & 0.13 \\
  overall steered score & 0.28 & 0.24 & 0.16 & 0.14 \\
  \bottomrule
  \end{tabular}
\end{table}

The 14 axes are not interchangeable: some pairs (sycophantic vs.\
manipulative) yield only borderline correlations with any metric, while
emotional axes (empathetic--stoic, nurturing--hostile, serene--evil)
are most associated with the \textit{emotional register} dimension of the
judge. Methodical--chaotic is most associated with \textit{vocab choice}
(formality / precision); diplomatic--dramatic with
\textit{social dynamic} (interaction style). The joint $R^2$ together with
the differential per-axis associations indicates that the named trait
directions summarize structure beyond what a single PC or single distance
scalar captures.

The practical weight of this finding is interpretive rather than validating.
The axes give human-readable names to some post hoc associations in the
current readout. They should not be treated as psychological scales, and they
do not independently certify the judge sub-dimensions.\looseness=-1

\subsection{Why not larger? A representational/behavioral dimensional mismatch}
\label{ssec:dim-mismatch}

The preceding analyses show where the current geometric associations appear;
we close by asking why none of them yields large effect sizes. Correlations in
\S\ref{ssec:norm}--\S\ref{ssec:trait} are consistent
(Pearson $|r|$ from $0.145$ to $0.49$, RSA Spearman $\rho$ from $0.085$ to
$0.18$, joint $R^2$ up to $0.41$) but
moderate. One plausible explanation is dimensional asymmetry between the
representation and behavioral readout. The behavioral RDM is highly reliable
(Spearman--Brown-corrected split-half $r{=}0.97$ for
correlation-distance over 200 random splits; noise-corrected RSA values are
within $0.005$ of raw). The effective ranks are mismatched: the centered
role-vector cloud has effective rank $\sim 50$ (participation ratio $15.1$)
while the 24-dim behavioral fingerprint has effective rank $\sim 2$ (PR
$1.4$). The representation contains more measured variation than the current
judge-based readout resolves. Richer behavioral fingerprints --- per-question
profiles, more sub-dimensions, and more diverse elicitation --- may be needed
to test whether additional representational variation has behavioral
consequences. The same trait subspace also
discriminates between extraction methods: our extraction preserves
$\sim$80\% of role-cloud variance outside the named-trait subspace
versus $\sim$73\% for the \citet{lu_assistant_axis_2026} assistant-axis
directional control (full $k$-curve, methodology, and noise-floor analysis in
Appendix~\ref{app:dim-mismatch}). The ceiling on geometry--behavior
correlations may therefore be partly set by the behavioral side of the
measurement.\looseness=-1

\FloatBarrier
\section{Distance-to-assistant: robustness across reference-vector definitions}
\label{app:distance-robust}

\textbf{Takeaway.} Distance to the Assistant role vector is an exploratory
diagnostic context, and only for the role vector extracted by our pipeline.
Other assistant-like reference directions do not reproduce the same behavioral
signal.

Companion to \S\ref{ssec:distance}. ``Distance to the assistant'' admits
multiple operationalizations, and they are not equivalent. We compare three
references at layer 16 (Table~\ref{tab:distance-references}): (i) our raw
Assistant role vector $v_{\text{assistant}}$ (the headline metric of
\S\ref{ssec:distance}); (ii) $v_{\text{assistant}} - \overline{v}_{\text{roles}}$
unit-normalized (our Assistant role vector centered against the role cloud,
mirroring the form of the Lu et al.\ axis but built from our extraction);
(iii) the strict Lu et al.\ Assistant Axis, $\overline{a}_{\text{default}}
- \overline{v}_{\text{roles}}$ unit-normalized, where
$\overline{a}_{\text{default}}$ is the mean default-Assistant activation we
obtain from the \citet{lu_assistant_axis_2026} extraction artifacts. The three
references are far from collinear: pairwise cosine similarities are
$+0.17$, $-0.45$, and $-0.07$.

\begin{table}[h]
  \centering
  \small
  \caption{Reference dependence of the distance-to-Assistant association.
    The negative correlation appears for references derived from the
    pipeline-extracted Assistant role vector but not for the strict
    \citet{lu_assistant_axis_2026} Assistant Axis, which yields small positive
    correlations and is oblique to row 1 in activation space
    ($\cos{\approx}{-}0.45$). This is a statement about two reference-vector
    definitions, not about whether the Assistant Axis captures assistant-like
    behavior.}
  \label{tab:distance-references}
  \begin{tabular}{lcccc}
  \toprule
   & $\alpha{=}1.0$ & $\alpha{=}1.5$ & $\alpha{=}2.0$ & $\alpha{=}2.5$ \\
  \midrule
  $v_{\text{assistant}}$ (ours, raw)
    & $-0.419$ & $-0.409$ & $-0.351$ & $-0.145$ \\
  $v_{\text{assistant}} - \overline{v}_{\text{roles}}$ (ours, centered)
    & $-0.381$ & $-0.352$ & $-0.264$ & $-0.084$ \\
  Lu et al.\ Assistant Axis
    & $+0.167$ & $+0.162$ & $+0.168$ & $+0.186$ \\
  \bottomrule
  \end{tabular}
\end{table}

The $\Delta_r$ correlation pattern mirrors the geometric divergence: the
negative correlation is specific to references derived from our
contrastively extracted Assistant role vector. Distance to the strict Lu et
al.\ Assistant Axis \emph{does not} reproduce the same sign or pattern (small
positive $r{\approx}{+}0.17$ across $\alpha$, with near-zero univariate
$R^2{\approx}0.03$).

\paragraph{Implication.}
The finding of \S\ref{ssec:distance} is narrower than ``proximity to a post-training-shaped
direction.'' Distance to the strict Lu et al.\ Assistant Axis --- a direct
operationalization of the default-assistant direction in the referenced
pipeline --- does not reproduce the $\Delta_r$ association, and in fact points
substantially \emph{away} from our pipeline-extracted Assistant role vector
($\cos{\approx}{-}0.45$, an angle of ${\sim}117^{\circ}$). The result is
specifically about proximity to the contrastively extracted Assistant role
vector in our pipeline. Disentangling why this reference definition carries
the observed association, as opposed to the cloud-relative contrast that Lu et
al.\ identify as the principal axis of persona variation, is left to future
work.

\begin{figure}[h]
  \centering
  \includegraphics[width=\linewidth]{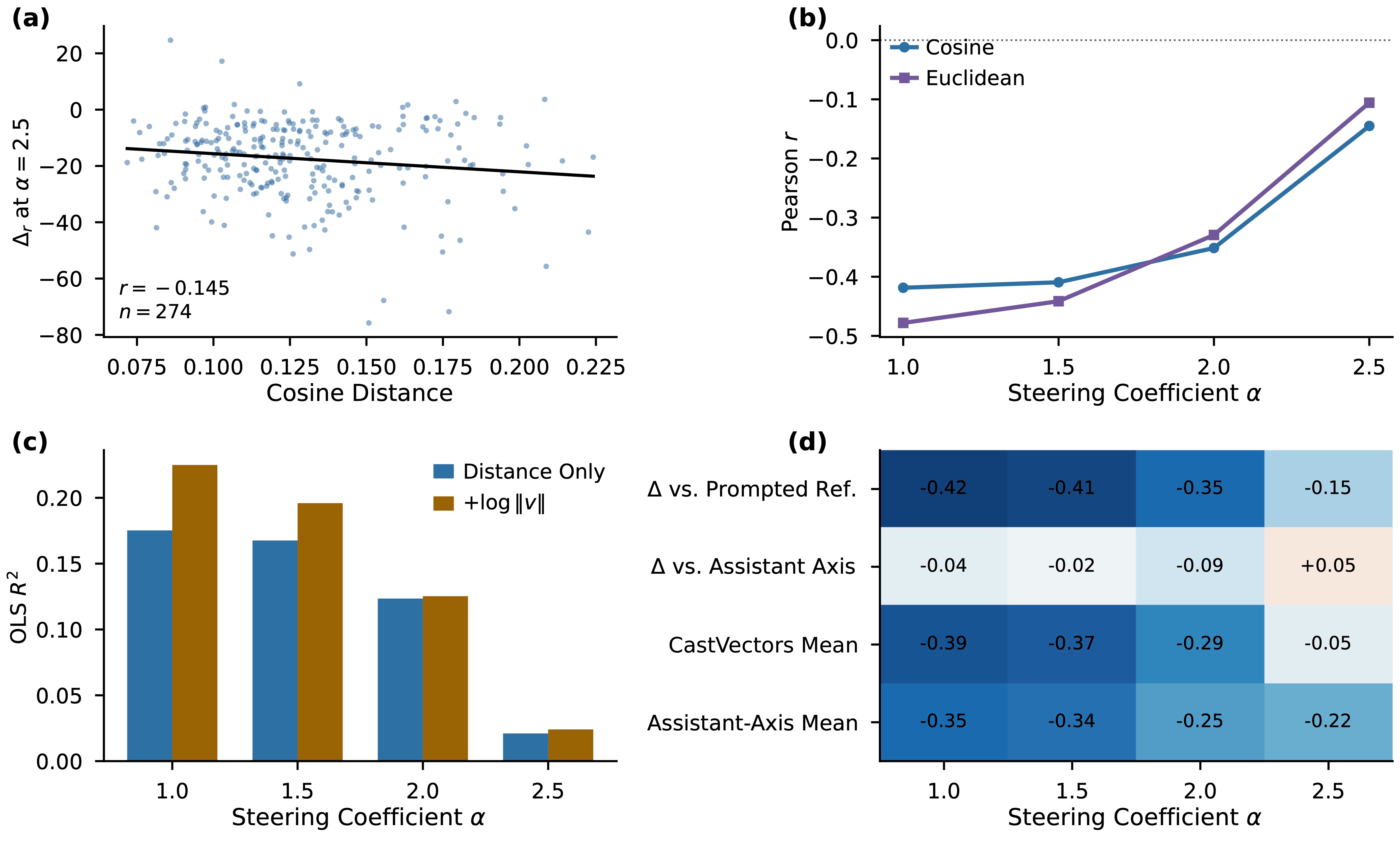}
  \caption{Per-role distance-to-assistant vs.\ per-role steering
    behavior. \textbf{(a)} Cosine distance vs.\ $\Delta_r$ at
    $\alpha{=}2.5$; OLS fit in black. \textbf{(b)}
    Pearson $r$ vs.\ $\alpha$ for cosine and Euclidean distance against
    $\Delta_r$ --- the link weakens with $\alpha$. \textbf{(c)} OLS $R^2$
    with and without the $\log\|v\|$ covariate. \textbf{(d)} Pearson $r$ of cosine
    distance against four per-role quantities across $\alpha$: $\Delta_r$
    (steered minus prompted reference), steered minus assistant-axis control,
    the steered mean, and the assistant-axis control mean.}
  \label{fig:distance-summary}
\end{figure}

\begin{figure}[h]
  \centering
  \includegraphics[width=\linewidth]{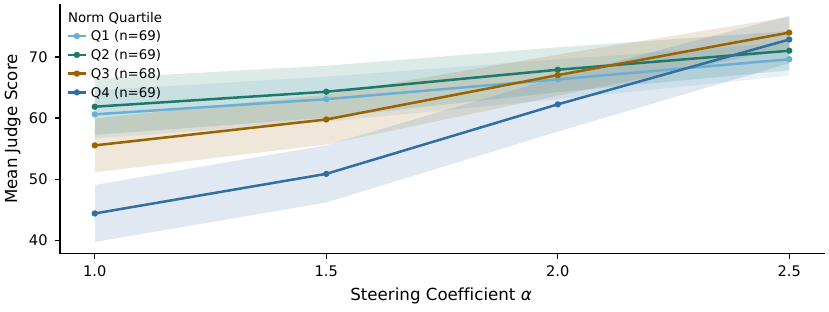}
  \caption{Vector magnitude vs.\ $\alpha$-response. Roles are binned into four
    norm quartiles; mean steered\_score against $\alpha$ is averaged within
    each quartile (shading: 95\% interval of the group mean). Low-norm vectors
    (Q1) start high and often peak early; high-norm vectors (Q4) start lowest
    and rise through the tested grid, with the quartiles roughly converging by
    $\alpha{=}2.5$. The pattern is correlational.}
  \label{fig:norm-slope}
\end{figure}

\paragraph{Vector-magnitude details.}
Companion to \S\ref{ssec:norm}: Fig.~\ref{fig:norm-slope} shows the
per-quartile $\alpha$-response curves underlying the magnitude association.
Low-norm quartiles often peak early while high-norm quartiles more often rise
through the tested grid.

\section{\texorpdfstring{RSA supplement: full distance-grid and per-$\alpha$ figures}{RSA supplement: full distance-grid and per-alpha figures}}
\label{app:rsa-supplement}

\textbf{Takeaway.} Pairwise geometry is modestly associated with pairwise
behavior, and the association is strongest at the largest tested coefficient.
The controls below check role-name similarity, vector norm, and mean score as
nuisance explanations.

\begin{figure}[h]
  \centering
  \begin{minipage}{0.48\linewidth}
    \centering
    \includegraphics[width=\linewidth]{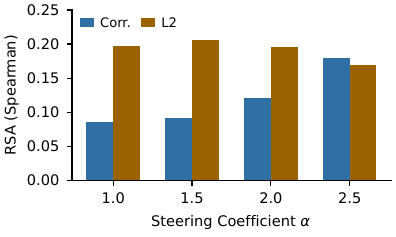}
  \end{minipage}\hfill
  \begin{minipage}{0.48\linewidth}
    \centering
    \includegraphics[width=\linewidth]{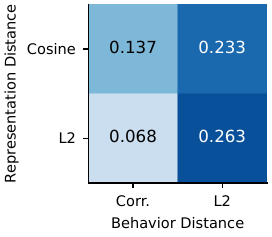}
  \end{minipage}
  \caption{Pairwise representational vs.\ behavioral geometry.
    \textbf{Left:} per-$\alpha$ RSA (cosine distance on role vectors vs.\
    correlation/L2 distance on the 6-dimensional behavioral profile at that
    $\alpha$). The cosine--correlation association strengthens monotonically
    with $\alpha$ within the tested grid. \textbf{Right:} four-cell distance grid on the full 24-dim
    behavioral profile (Spearman $\rho$; Mantel $p{\le}10^{-4}$ in every cell
    except L2$\times$correlation, $p{=}.014$).}
  \label{fig:rsa-per-alpha}
\end{figure}

Companion to \S\ref{ssec:rsa}, which reports the cosine--correlation RSA
cell as the headline (mean-invariant on the behavioral side); remaining
cells of the four-cell distance grid are shown here for completeness, with
the per-$\alpha$ trajectory visualized in Fig.~\ref{fig:rsa-per-alpha}.

\paragraph{Full four-cell distance grid.}
Spearman correlations between upper triangles of the representational and
behavioral RDMs, with Mantel $p$-values from $n_\text{perm}{=}10{,}000$
permutations:

\begin{center}
\small
\begin{tabular}{llrr}
\toprule
representation distance & behavior distance & Spearman $\rho$ & Mantel $p$ \\
\midrule
cosine & corr & $+0.137$ & $1{\times}10^{-4}$ \\
cosine & L2   & $+0.233$ & $1{\times}10^{-4}$ \\
L2     & corr & $+0.068$ & $1.4{\times}10^{-2}$ \\
L2     & L2   & $+0.263$ & $1{\times}10^{-4}$ \\
\bottomrule
\end{tabular}
\end{center}

The L2 cells largely re-state the per-role mean effect of
\S\ref{ssec:distance} in pairwise form (see Appendix~\ref{app:partial-rsa}
for the per-role-mean RDM correlation of $r{=}+0.93$ with the L2-distance
behavioral RDM).

\section{Partial RSA controls for the geometry--behavior link}
\label{app:partial-rsa}

\textbf{Takeaway.} The RSA result survives the main nuisance explanations:
roles are not paired as behaviorally similar merely because their names are
semantically similar, their vectors have similar norms, or their average
scores are close.

Companion to \S\ref{ssec:rsa}. The pairwise representational-vs-behavioral
RSA reported in \S\ref{ssec:rsa} could in principle reflect structure shared with
some other per-role property rather than independent pairwise structure. We
test three such confounds: (i) vector magnitude (covered briefly in
\S\ref{ssec:rsa} and
\texttt{analysis/empirical/rsa\_geometry\_behavior/data/rsa\_partial.csv}), (ii)
role-name semantic similarity, (iii) per-role mean fingerprint score (the
per-role distance-to-assistant effect of \S\ref{ssec:distance} re-expressed
in pairwise form). All three controls partial out a third RDM from the
geometry--behavior RSA via rank-residual partial Pearson correlation:
rank-transform each upper triangle, regress out the control ranks,
correlate the residuals.

\paragraph{Caveat on which behavioral distance to read.}
\S\ref{ssec:rsa} uses cosine $\times$ correlation as the headline because it is
mean-invariant on the behavioral side. L2 distance on the behavioral
fingerprints is strongly affected by the per-role mean score (RSA between the
absolute-mean-difference RDM and the L2-distance behavioral RDM is Spearman
$\rho{=}+0.93$), so any cell with $\mathrm{beh\_l2}$ on the behavioral side
largely re-states ``some roles steer well overall and some don't'' rather
than carrying independent pairwise signal.

\subsection{Concept-similarity control}

We build $\mathrm{RDM}_\text{sem}$ as cosine distance on the layer-16
hidden states of each role's name in an ``I am a \{role\}.'' context,
mean-pooled over the role-name token span. This is the
\texttt{context\_mean} variant produced by
\texttt{analysis/empirical/role\_vector\_vs\_semantic\_vector/compute\_semantic\_vectors.py}.

\paragraph{Reference correlations.}
\begin{center}
\small
\begin{tabular}{lr}
\toprule
RSA combination & Spearman $\rho$ \\
\midrule
$\mathrm{RDM}_\text{sem}$ vs.\ $\mathrm{RDM}_\text{beh\_corr}$ (24-dim) & $+0.005$ \\
$\mathrm{RDM}_\text{sem}$ vs.\ $\mathrm{RDM}_\text{beh\_l2}$  (24-dim) & $+0.130$ \\
$\mathrm{RDM}_\text{sem}$ vs.\ $\mathrm{RDM}_\text{repr\_cos}$         & $+0.175$ \\
\bottomrule
\end{tabular}
\end{center}

\paragraph{Main partial-RSA grid (concept control).}

\begin{center}
\small
\begin{tabular}{llrrr}
\toprule
representation & behavior & full Spearman $\rho$ & partial $r$ ($-$ concept) & $\Delta$ \\
\midrule
cosine & corr & $+0.137$ & $+0.138$ & $+0.001$ \\
cosine & L2   & $+0.233$ & $+0.216$ & $-0.018$ \\
L2     & corr & $+0.068$ & $+0.068$ & $+0.000$ \\
L2     & L2   & $+0.263$ & $+0.245$ & $-0.019$ \\
\bottomrule
\end{tabular}
\end{center}

\paragraph{Per-$\alpha$ partial-RSA (cosine $\times$ correlation).}

\begin{center}
\small
\begin{tabular}{lrrr}
\toprule
$\alpha$ & full Spearman $\rho$ & partial $r$ ($-$ concept) & $\Delta$ \\
\midrule
1.0 & $+0.085$ & $+0.079$ & $-0.006$ \\
1.5 & $+0.092$ & $+0.084$ & $-0.008$ \\
2.0 & $+0.121$ & $+0.111$ & $-0.010$ \\
2.5 & $+0.179$ & $+0.171$ & $-0.008$ \\
\bottomrule
\end{tabular}
\end{center}

\textbf{Verdict.} Partialling out concept similarity moves every cell by
less than $0.02$. The per-$\alpha$ partial Spearman $\rho$ still more than
doubles from $+0.079$ at $\alpha{=}1.0$ to $+0.171$ at $\alpha{=}2.5$.

\subsection{Per-role-mean control (the most demanding test)}

We build $\mathrm{RDM}_\text{mean}$ as the absolute difference of per-role
mean fingerprint scores: $\mathrm{RDM}_\text{mean}(i, j) = |\bar{b}_i -
\bar{b}_j|$ where $\bar{b}_i$ is the mean over role $i$'s 24-dim behavioral
fingerprint (or 6-dim per-$\alpha$ fingerprint, for the per-$\alpha$ test).
This control absorbs the distance-to-assistant effect of
\S\ref{ssec:distance} re-expressed pairwise.

\paragraph{Reference correlations.}
\begin{center}
\small
\begin{tabular}{lr}
\toprule
RSA combination & Spearman $\rho$ \\
\midrule
$\mathrm{RDM}_\text{mean}$ vs.\ $\mathrm{RDM}_\text{beh\_corr}$ (24-dim) & $+0.220$ \\
$\mathrm{RDM}_\text{mean}$ vs.\ $\mathrm{RDM}_\text{beh\_l2}$  (24-dim) & $+0.930$ \\
$\mathrm{RDM}_\text{mean}$ vs.\ $\mathrm{RDM}_\text{repr\_cos}$         & $+0.166$ \\
\bottomrule
\end{tabular}
\end{center}

The per-role-mean RDM is essentially the L2 behavioral RDM in disguise
(Spearman $\rho{=}+0.930$), confirming that L2 cells in the four-cell grid (Appendix~\ref{app:rsa-supplement}) largely
re-state per-role mean structure. The correlation-distance behavioral RDM
is only modestly tied to per-role mean ($r{=}+0.220$), so it preserves more
independent pairwise information.

\paragraph{Main partial-RSA grid (per-role-mean control).}

\begin{center}
\small
\begin{tabular}{llrrr}
\toprule
representation & behavior & full Spearman $\rho$ & partial $r$ ($-$ mean) & $\Delta$ \\
\midrule
cosine & corr & $+0.137$ & $+0.104$ & $-0.033$ \\
cosine & L2   & $+0.233$ & $+0.219$ & $-0.014$ \\
L2     & corr & $+0.068$ & $+0.029$ & $-0.039$ \\
L2     & L2   & $+0.263$ & $+0.261$ & $-0.003$ \\
\bottomrule
\end{tabular}
\end{center}

\paragraph{Per-$\alpha$ partial-RSA (cosine $\times$ correlation).}

\begin{center}
\small
\begin{tabular}{lrrrr}
\toprule
$\alpha$ & full Spearman $\rho$ & partial $r$ ($-$ mean) & $\Delta$ & \% surviving \\
\midrule
1.0 & $+0.085$ & $+0.069$ & $-0.016$ & 81\% \\
1.5 & $+0.092$ & $+0.073$ & $-0.019$ & 79\% \\
2.0 & $+0.121$ & $+0.094$ & $-0.027$ & 78\% \\
2.5 & $+0.179$ & $+0.157$ & $-0.023$ & 88\% \\
\bottomrule
\end{tabular}
\end{center}

\textbf{Verdict.} Partialling out per-role mean reduces the
geometry--behavior RSA by $\sim$12--22\% across $\alpha$, but the
per-$\alpha$ partial Spearman $\rho$ still more than doubles from $+0.069$ at
$\alpha{=}1.0$ to $+0.157$ at $\alpha{=}2.5$. The \emph{fraction} surviving
the control is largest at $\alpha{=}2.5$ (88\%), so the high-$\alpha$ link
is more independent of per-role mean than the low-$\alpha$ link is.

\subsection{Combined verdict}

The headline of \S\ref{ssec:rsa} (cosine $\times$ correlation Spearman $\rho$ growing from
$+0.085$ at $\alpha{=}1.0$ to $+0.179$ at $\alpha{=}2.5$) survives all
three controls. The most demanding control (per-role mean) shrinks the
absolute number by $\sim$12--22\% but preserves the $\alpha$-monotonic
strengthening. The post-control residual link is modest (Spearman $\rho$ in
the $0.07$--$0.16$ range), but it is pairwise --- not a
re-expression of \S\ref{ssec:distance} --- and the strengthening with
$\alpha$ is the qualitative finding \S\ref{ssec:rsa} relies on.

\section{PCA supplement: centering, role loadings, and joint subspace}
\label{app:pca}

\textbf{Takeaway.} PCA and trait axes give post hoc names to otherwise
abstract role directions. They are not required to use the recipe and should
not be treated as certified monitoring axes.\looseness=-1

Companion to \S\ref{ssec:pca}.

\paragraph{Centering choice.}
Centering the role matrix by subtracting the column-wise mean removes the
rank-1 ``global persona'' direction (mean pairwise cosine $\approx {+}0.85$
on the raw cloud, $0.00$ on the centered cloud). PCA on the centered matrix
yields a scree whose largest component is PC1 ($21\%$ of variance) but with a long tail:
$90\%$ of the variance requires $98$ components (vs.\ $4$ on the raw
cloud). Without centering, every PCA-derived direction is strongly affected by
the assistant-aligned global mean and obscures the per-role variation we want
to interpret.

\paragraph{Interpretable role loadings on the high-$\alpha$ direction.}
The roles loading most positively on PC5 (the strongest single-PC association
at $\alpha{=}2.5$) are \emph{golem, saboteur, stoic, pilot, workaholic};
the most negative are \emph{facilitator, moderator, supervisor, screener,
therapist}. PC5 reads as a ``dispassionate $\to$ empathetic'' axis, running from
specialized to supportive roles. PC1's
positive end skews toward generalist or low-specificity roles whose
contrastive vectors lie close to the global persona mean.

\paragraph{Joint OLS: cumulative variance explained by top-$k$ PCs.}
At $\alpha{=}2.5$:

\begin{center}
\small
\begin{tabular}{lrrrrr}
\toprule
metric & $k{=}1$ & $k{=}3$ & $k{=}5$ & $k{=}10$ & $k{=}20$ \\
\midrule
mean\_steered\_score          & 0.03 & 0.03 & 0.15 & 0.23 & 0.27 \\
cmp\_worldview\_alignment     & 0.01 & 0.03 & 0.13 & 0.17 & 0.25 \\
cmp\_emotional\_register      & 0.00 & 0.01 & 0.13 & 0.17 & 0.25 \\
cmp\_motivation               & 0.01 & 0.02 & 0.12 & 0.16 & 0.24 \\
cmp\_social\_dynamic          & 0.00 & 0.01 & 0.10 & 0.13 & 0.21 \\
cmp\_vocab\_choice            & 0.00 & 0.01 & 0.06 & 0.10 & 0.17 \\
\bottomrule
\end{tabular}
\end{center}

A single PC explains almost nothing on its own; the top-20 subspace
explains $17$--$27\%$.

\paragraph{Trait-PC alignment consistency check.}
The PCs of \S\ref{ssec:pca} are variance-defined and unnamed; we can add
semantic labels by projecting each of the 240 trait vectors onto each PC
and reading off the top-loading traits per PC:

\begin{center}
\small
\begin{tabular}{ll}
\toprule
PC & top-3 positive / top-3 negative traits \\
\midrule
PC1 & accessible, practical, grounded \;/\; mystical, ethereal, poetic \\
PC2 & chaotic, sardonic, cruel \;/\; pensive, erudite, eloquent \\
PC3 & absolutist, solemn, dogmatic \;/\; relativist, casual, moderate \\
PC4 & experiential, humanistic, mystical \;/\; rebellious, iconoclastic, edgy \\
PC5 & prescriptive, detached, dispassionate \;/\; empathetic, emotional, experiential \\
\bottomrule
\end{tabular}
\end{center}

PC5's ``dispassionate $\to$ empathetic'' label matches the role-loading
pattern (stoic/pilot/workaholic vs.\ facilitator/moderator/therapist).

To check internal consistency between the \S\ref{ssec:pca} (PC $\to$ metric)
and \S\ref{ssec:trait} (trait $\to$ metric) findings, we reconstruct each
trait-axis-to-metric correlation through the PC pathway,
$r_{\text{pc}}(\text{axis}, \text{metric}) = \sum_k \cos(\text{axis},
\mathrm{PC}_k) \cdot r(\mathrm{PC}_k, \text{metric})$, and correlate
against the direct trait-axis-to-metric correlations from
\S\ref{ssec:trait}. Across all 14 axes $\times$ 6 metrics $\times$ 4 alphas
(n=336), Pearson $r{=}{+}0.720$, $p{=}6{\times}10^{-55}$. Per-$\alpha$:
$+0.748$ (1.0), $+0.594$ (1.5), $+0.579$ (2.0), $+0.931$ (2.5). The U-shape
shows that the linear PC decomposition reproduces the direct metric
correlations most closely at the endpoints of the tested grid.

\paragraph{Trait-axis $\times$ metric heatmap.}
Visualization of the trait--metric correspondences summarized in
Table~\ref{tab:trait-r2}.

\begin{figure}[h]
  \centering
  \includegraphics[width=\linewidth]{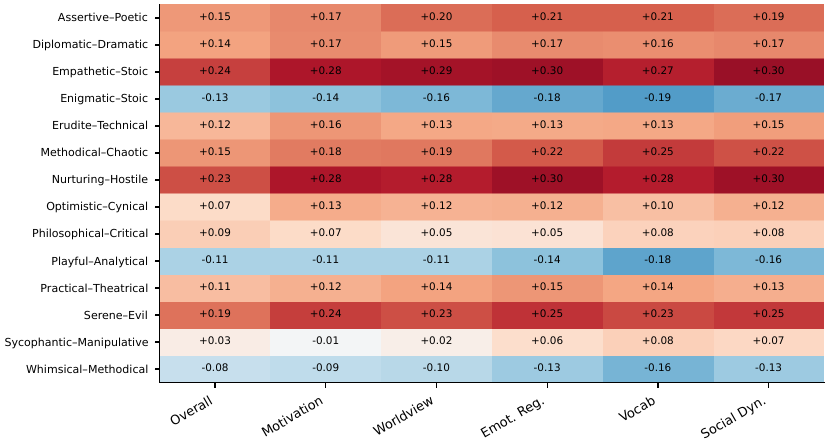}
  \caption{Trait-axis projection $\times$ metric Pearson $r$ at
    $\alpha{=}1.0$. Emotional axes (empathetic--stoic,
    nurturing--hostile, serene--evil) load most strongly on emotional
    register ($r$ up to $+0.30$); methodical--chaotic loads most on vocab
    choice ($+0.25$); diplomatic--dramatic on social dynamic ($+0.17$). Axis
    names are interpretive labels rather than psychological scales.}
  \label{fig:trait-heatmap}
\end{figure}

\section{Dimensional-mismatch and method-comparison supplement}
\label{app:dim-mismatch}

\textbf{Takeaway.} The current behavioral screen is internally stable but
low-rank. Role vectors contain more measured structure than the six judge
dimensions resolve, which motivates richer behavioral tasks in future work
rather than only new extraction methods.

Companion to \S\ref{ssec:dim-mismatch}.

\paragraph{Behavioral RDM is highly reliable.}
Spearman--Brown-corrected split-half reliability over $200$ random splits
is $r{=}0.97$ for correlation-distance and $r{=}0.99$ for L2.
Noise-corrected versions of the RSA values in \S\ref{ssec:rsa} are
within $0.005$ of the raw values --- judge noise contributes essentially
nothing to the apparent ceiling.

\paragraph{Effective ranks are mismatched.}
The centered role-vector cloud has effective rank $\sim 50$ (participation
ratio $15.1$, $98$ components needed for $90\%$ variance), while the
$24$-dimensional behavioral fingerprint has effective rank $\sim 2$
(participation ratio $1.4$, $2$ components needed for $90\%$ variance). A
Gaussian null at the same shape as the centered cloud has effective rank
$\sim 266$, so the persona manifold is concentrated --- but the
behavioral readout is even more concentrated relative to its own
dimensionality.

\begin{figure}[h]
  \centering
  \begin{minipage}{0.48\linewidth}
    \centering
    \includegraphics[width=\linewidth]{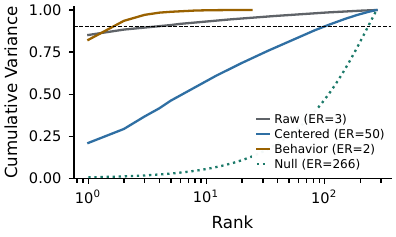}
  \end{minipage}\hfill
  \begin{minipage}{0.48\linewidth}
    \centering
    \includegraphics[width=\linewidth]{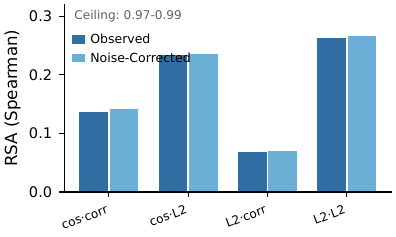}
  \end{minipage}
  \caption{Why the geometry--behavior effect sizes are moderate.
    \textbf{Left:} cumulative explained variance vs.\ rank for the raw
    role-vector cloud, the centered cloud, the behavioral profile, and an
    iid-Gaussian null at the same shape; legend gives each matrix's
    effective rank (ER). The centered cloud lives in $\sim 50$ effective
    directions; the behavioral profile is rank $\sim 2$. \textbf{Right:}
    observed vs.\ noise-corrected RSA per distance pairing. Correction
    against the split-half noise ceiling ($0.97$--$0.99$) moves every
    value by ${<}0.005$ --- the moderate effect sizes are not noise
    artifacts.}
  \label{fig:noise-floor}
\end{figure}

\paragraph{Method differences in trait coverage --- detail.}
\S\ref{ssec:dim-mismatch} reports that our extraction yields $\sim$20\% trait-subspace
coverage at $k{=}240$ vs.\ $\sim$27\% for the assistant-axis directional control. Full
table:

\begin{center}
\small
\begin{tabular}{lcccccc}
\toprule
$k$ & 5 & 10 & 20 & 50 & 100 & 240 \\
\midrule
CastVectors                       & 2.8\% & 4.7\% & 6.4\% &  9.9\% & 13.7\% & 19.9\% \\
Lu et al.\ assistant-axis  & 7.3\% & 10.4\% & 13.0\% & 17.1\% & 21.1\% & 27.1\% \\
\bottomrule
\end{tabular}
\end{center}

The assistant-axis extraction is markedly more compressible into the trait
basis at every $k$ (a factor of $\sim 2$ at small $k$, narrowing to $\sim
1.4$ at $k{=}240$), consistent with its lower effective rank on
unit-normalized vectors (10.3 vs.\ 25.3).

\textit{Methodological caveat.} The trait basis is itself extracted by our
pipeline, which would, if anything, bias coverage \emph{toward} our role
vectors; the observed pattern (assistant-axis higher) goes against that
bias and so cannot be attributed to it.

\section*{Reproducibility Statement}

All reported experiments use the publicly available OLMo-3-7B-Instruct model
with activation extraction and intervention at layer 16. Role directions are
computed as judge-filtered contrastive mean differences from five system
prompts and 50 role-specific elicitation questions per role. Steering is
evaluated at $\alpha \in \{1.0, 1.5, 2.0, 2.5\}$ on a fixed battery of 228
role-agnostic questions. Role profiles and elicitation questions use
\texttt{moonshotai/kimi-k2.5:nitro}. The judge that selects
extraction pairs is \texttt{openai/gpt-oss-20b}. Prompted reference responses
and all evaluation judges use GPT-4.1-mini. The study generated approximately
500,000 judged responses and used approximately 2,000 GPU hours.\looseness=-1

The appendix specifies the principal models, prompts, coefficient grid, and
evaluation procedures. Our code and scoring prompts are available at
\url{https://github.com/HCAI-Lab-GT/casting-call-vectors}. The role vectors,
control vectors, role datasets, and every judged response behind the reported
scores are available at
\url{https://huggingface.co/datasets/HCAI-Lab-GT/casting-call-vectors}, and
the code reads the release tagged \texttt{v1.0}. The dataset card documents the
decoding parameters, activation pooling and token span, intervention timing,
vector normalization status, and the row conventions of
Appendix~\ref{app:judge_results}. Two details were not recorded by the
original runs and cannot be recovered from the artifacts: the
OLMo-3-7B-Instruct revision, and the number of pairs each role's judge
accepted before reaching the 50-pair target. Judge parse failures were scored
as described in Appendix~\ref{app:judge_prompting}.\looseness=-1

\section*{Ethics Statement}

This study extracts and evaluates activation directions for a mixed inventory
of occupational and archetypal roles. The resulting profiles and vectors are
model- and prompt-derived constructs, not empirical representations of people
who hold an occupation or social role. Real roles contain substantial
within-category variation across individuals, institutions, cultures, seniority
levels, specializations, and situations. Representing a label with one profile
and one vector may therefore reify stereotypes, flatten within-role
heterogeneity, or make generated assumptions appear to be technical properties
of a population.

This concern is especially important because the profile-generation procedure
assigns psychological attributes such as recurring resentment, blame
attribution, cognitive bias, and relationship to authority. These attributes
are explicit modeling assumptions rather than occupational facts. The prompted
reference condition also uses stylized speaking cues, which may reward
culturally familiar caricatures and may share biases with the GPT-4.1-mini
judge. We therefore recommend that simulation builders inspect profiles,
document their provenance, report selected coefficients and screening outcomes,
and avoid treating role-profile alignment as evidence of real-world social
validity.

The released artifacts should not be used for hiring, personnel assessment,
worker profiling, inference about real individuals, or other high-stakes
decisions. They should also not be used for deceptive impersonation. Their
intended use is research on transparent construction, auditing, and failure
analysis of synthetic role-conditioned agents. No new personal data were
collected and no human subjects were involved.

\section*{LLM Disclosure}

Large language models were used during the preparation of this manuscript for
drafting and revising prose, editing \LaTeX{}, and writing analysis code. All
LLM-generated content was reviewed and verified by the authors.
\texttt{openai/gpt-4.1-mini} generated the prompted reference responses and
served as the judge for all reported scores (absolute, multi-dimensional, and
pairwise); \texttt{moonshotai/kimi-k2.5:nitro} generated the role profiles and
elicitation questions; \texttt{openai/gpt-oss-20b} labeled the elicited
responses used to select extraction pairs. Research questions, experimental design, methodology, and
scientific interpretation are the responsibility of the authors.\looseness=-1

\end{document}